\documentclass[gmd, manuscript]{copernicus}
  
\begin{document}

\title{Surrogate-assisted Bayesian inversion for landscape and basin evolution models }


\Author[1,2]{Rohitash}{Chandra}
\Author[2]{Danial}{Azam}
\Author[3]{Arpit}{Kapoor} 
\Author[2]{R. Dietmar}{Müller}

\affil[1]{School of Mathematics and Statistics, University 
of New South Wales, Sydney, NSW 2052, Australia}
\affil[2]{EarthByte Group, School of Geosciences, University of 
Sydney,  Sydney, NSW 2006, Australia}

\affil[3]{ Department of Computer Science and Engineering,
SRM Institute of Science and Technology, Tamil Nadu, India}

\runningtitle{TEXT}

\runningauthor{TEXT}

\correspondence{Rohitash Chandra (rohitash.chandra@unsw.edu.au)}

\received{}
\pubdiscuss{} 
\revised{}
\accepted{}
\published{}

\firstpage{1}

\maketitle

\begin{abstract}

The complex and computationally expensive nature of   landscape  evolution models pose   significant challenges in the inference and optimisation of unknown model parameters. Bayesian inference provides a methodology for estimation and uncertainty quantification of unknown model parameters. In our previous work, we developed parallel tempering Bayeslands as a framework for parameter estimation and uncertainty quantification for the Badlands landscape evolution model.   Parallel tempering Bayeslands features high-performance computing that can feature dozens of processing cores running in parallel to enhance computational efficiency.    Nevertheless, the procedure remains computationally challenging since thousands of samples need to be drawn and evaluated.  In large-scale landscape evolution problems, a single model evaluation can take from several minutes to hours and in some instances, even days or weeks.   Surrogate-assisted optimisation has been  used for several computationally expensive engineering problems which motivate its use in optimisation and inference  of  complex geoscientific models. The use of surrogate models can speed up parallel tempering Bayeslands by developing computationally inexpensive models to mimic expensive ones.   In this paper,  we apply surrogate-assisted parallel tempering where the surrogate mimics a landscape evolution model   by estimating the likelihood function from the model.  We employ a neural network-based surrogate model that learns from the history of samples generated.  The entire framework is developed in a parallel computing infrastructure to take advantage of parallelism. The results show that the proposed methodology is effective in lowering the computational cost significantly while retaining the quality of model predictions.

\end{abstract}

\copyrightstatement{}

\introduction   
  The Bayesian methodology provides a probabilistic approach for the estimation of unknown parameters in complex models \citep{sambridge1999geophysical,neal1996sampling,chandra2019langevin}. We can view a  deterministic geophysical forward model as a probabilistic model via Bayesian inference which is also known as Bayesian inversion which has been used for landscape evolution \citep{Chandra2018_Bayeslands_,chandra2018PT-Bayes_}, geological reef evolution models \citep{JPall_BayesReef2020} and other geoscientific models \citep{sambridge1999geophysical,sambridge2013parallel,scalzo2019efficiency,OLIEROOK2020}. Markov Chain Monte Carlo (MCMC) sampling is typically used to implement Bayesian inference that involves the estimation and uncertainty quantification of unknown parameters  \citep{hastings1970monte,metropolis1953equation, neal2012bayesian,neal1996sampling}.  Parallel tempering MCMC  \citep{marinari1992simulated,geyer1995annealing}     features multiple replicas to provide    a balance between  exploration and exploitation   which makes them suitable for irregular and multi-modal distributions \citep{patriksson2008temperature,hukushima1996exchange}. In contrast to canonical sampling methods, we can implement  parallel tempering more easily in a  parallel computing architecture \citep{lamport1986interprocess}.

Our previous work presented parallel tempering Bayeslands for parameter estimation and uncertainty quantification for landscape evolution models (LEMs)  \citep{chandra2018PT-Bayes_}.   Parallel tempering Bayeslands features parallel computing to enhance  computational efficiency of inference for the Badlands LEM.  Although we used parallel computing, the procedure was computationally challenging since thousands of samples were drawn and evaluated \citep{chandra2018PT-Bayes_}. In large-scale LEMs, running a single model can take several  hours, to days or weeks and usually thousands of model runs are required for inference of unknown model parameters. Hence, it is important to enhance parallel tempering Bayeslands which can also be applicable    for other complex geoscientific models. One of the ways to address this problem is through surrogate-assisted estimation.

Surrogate assistant optimisation refers to the use of statistical and machine learning models  for developing approximate simulation or surrogate of the actual model \citep{jin2011surrogate}.  Since typically  optimisation methods lack a rigorous approach for uncertainty quantification,  Bayesian inversion becomes as an alternative choice particularly for complex geophysical numerical models \citep{sambridge2013parallel,sambridge1999geophysical}. The major advantage of a surrogate model is its computational efficiency when compared to the equivalent numerical physical forward model   \citep{ong2003evolutionary,zhou2007combining}. In the optimization literature, surrogate utilization is also known as response surface methodology \citep{Douglas1977,letsinger1996response},  and applicable for a wide range of  engineering problems   \citep{tandjiria2000reliability,ong2005surrogate} such as aerodynamic wing  design \citep{ong2003evolutionary}. Several approaches have been used to improve the way surrogates are utilised. 
 \citep{zhou2007combining} combined global and local surrogate models to accelerate evolutionary optimisation.   
 \citep{lim2010generalizing} presented a generalised surrogate-assisted evolutionary computation framework to unify diverse surrogate models during optimisation and taking into account uncertainty in estimation. Jin \citep{jin2011surrogate} reviewed a range of problems such as single, multi-objective,  dynamic, constrained, and multi-modal optimisation problems \citep{diaz2016review}.  In the Earth sciences, examples for surrogate assisted approaches include modeling  water resources \citep{razavi2012review,asher2015review}, atmospheric general circulation models  \citep{Scher2018}, computational oceanography \citep{van2007fast}, carbon-dioxide (CO2) storage and oil recovery   \citep{ampomah2017co}, and debris flow models \citep{navarro2018surrogate}.

Given that Bayeslands is implemented using parallel computing, the challenge is in implementing surrogates across different processing cores. Recently, we developed surrogate-assisted parallel tempering has for Bayesian neural networks, which used a global-local surrogate framework to execute surrogate training in the master processing core that manages the replicas running in parallel \citep{chandra2020SAPT}.  The global surrogate refers to the main surrogate model that features training data combined from different replicas running in parallel cores. Local surrogate model refers to the surrogate model in the given replica that incorporates knowledge from the global surrogate to make a prediction given new input parameters. Note that the training only takes place  in the global surrogate and the prediction or estimation for pseudo-likelihood only takes place in the local surrogates.  The method gives promising results where prediction performance is maintained while lowering computational time using surrogates.

In this paper, we present an application of surrogate-assisted parallel tempering \citep{chandra2020SAPT} for Bayesian inversion of LEMs using  parallel computing infrastructure. We use the Badlands LEM model  \citep{salles2018pybadlands}  as a case study to demonstrate the framework. Overall, the framework features the surrogate-model which mimics the Badlands model and estimates the likelihood function to evaluate the proposed parameters.  We employ a neural network model as the surrogate that learns from the history of samples from the parallel tempering MCMC.   We apply the method to several selected benchmark landscape evolution and sediment transport/deposition problems and show the quality of the estimation of the likelihood given by the surrogate when compared to the actual Badlands model. 

\section{Background and Related Work}
 
\subsection{Bayesian inference } 

  Bayesian inference is typically implemented by employing MCMC sampling methods   that update the probability for a hypothesis as more information becomes available. The hypothesis is given by a prior probability distribution (also known as the prior) that expresses one's belief about a quantity (or free parameter in a model) before some data is taken into account. Therefore, MCMC methods provide a probabilistic approach for estimation of free parameters  in a wide range of  models  \citep{raftery1996,van2016simple}. The likelihood function is a way to evaluate the sampled parameters for a model with given observed data. In order to evaluate the likelihood function, one would need to run the given model, which in our case is the Badlands model. The likelihood function is used with the Metropolis-criteria to either accept or reject a proposal. When accepted, the proposal becomes part of the posterior distribution, which essentially provides the estimation of the free parameter with uncertainties. The sampling process is iterative and requires thousands of samples are drawn until convergence. In our case, convergence is defined by a predefined number of samples or until the likelihood function has reached a specific value.

\subsection{Badlands model and Bayeslands framework }

  LEMs incorporate different driving forces such as tectonics or climate variability \citep{Whipple2002,Tucker10,salles2018pybadlands,Campforts2017,Adams2017} and  combine empirical data and conceptual methods into a set of mathematical equations. \textit{Badlands} (basin and landscape dynamics) \citep{salles2018pybadlands,salles2016badlands} is an example of such a model that can be used to reconstruct landscape evolution and associated sediment fluxes \citep{Howard1994,Hobley2011}.   \textit{Badlands} LEM model \citep{salles2018pybadlands}  simulates landscape evolution and sediment transport/deposition with given  parameters such as the \textit{precipitation} rate  and rock \textit{erodibility} coefficient.  The Badlands  LEM simulates landscape dynamics which requires an initial topography exposed to climate and geological factors over time.

Bayeslands essentially provides the estimation of unknown Badlands parameters with Bayesian inference via MCMC sampling \citep{chandra2018PT-Bayes_}.  We use the final or present-day topography at time $T$  and expected sediment deposits at selected intervals to evaluate the quality of proposals during sampling.   In this way, we constrain the set of unknown parameters  ($\theta$) using ground-truth data ($\mathbf D$).  The prior distribution (also known as prior) refers to one's belief in the 
distribution of the parameter without taking into account  the evidence  or   
data.    Bayeslands   estimates   $\theta$  so that the simulated topography by Badlands can resemble the ground-truth topography $\mathbf D$ to some degree.  Bayeslands samples  the posterior distribution $p(\theta|\mathbf
D)$ using  principles of  Bayes rule
 \[
p(\theta|\mathbf D)=\frac{p(\mathbf D|\theta)p(\theta)}{P(\mathbf D)}
\]
where, $p(\mathbf D|\theta)$ is the likelihood of the data given the parameters,  
$p(\theta)$ is the prior, and $p(\mathbf D)$ is a normalizing constant and equal to 
$\int p(\mathbf D|\theta)p(\theta)d\theta$.     We note that the prior ratio cancels out since    we use a uniform distribution for the priors.

\section{Methodology}

\subsection{Benchmark landscape evolution problems}

 \begin{figure}[htbp!]
  \begin{center}   
  \includegraphics[width=150mm]{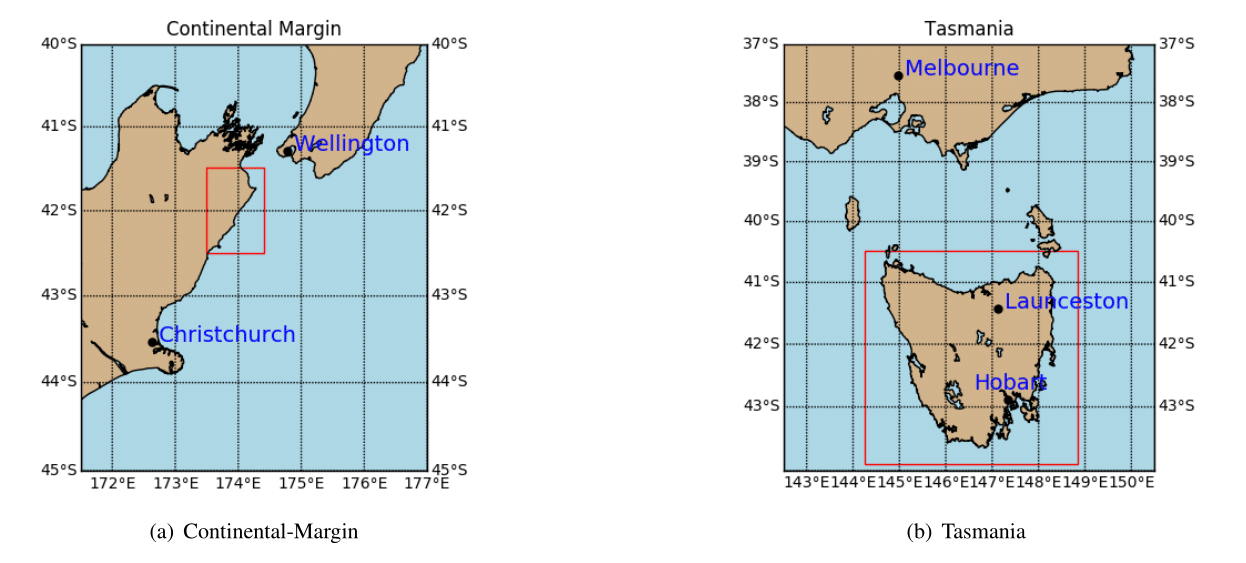}
    \caption{Location of (a) Continental-Margin problem shown taken from South Island of New Zealand 
    (b) Tasmania, Australia  with latitude and longitude information  shown  in degrees. }
 \label{fig:cm-map}
  \end{center}
\end{figure}

 We select two benchmark landscape problems from parallel tempering Bayeslands \citep{chandra2018PT-Bayes_} that are adapted from earlier work \citep{Chandra2018_Bayeslands_}.  These include    \textit{Continental Margin} (CM)  and   \textit{Synthetic-Mountain} (SM) which are chosen due to the computational time taking for running a single model since  they use less than five seconds to run a single model on a single central processing unit (CPU).  These problems are well suited for a parameter evaluation for the proposed surrogate-assisted Bayesian inversion framework. In order to demonstrate an application which is computationally expensive,  we introduce another problem, which features the landscape evolution of Tasmania in Australia for a million years that features the region shown in  Figure \ref{fig:cm-map}~Panel(b). The Synthetic-Mountain landscape evolution is a synthetic problem while the Continental-Margin problem is a real-world problem based on the topography of a region along the eastern margin of the South Island of New Zealand as shown in  Figure shown in  Figure \ref{fig:cm-map}~Panel(a).  We   use Badlands to evolve the initial landscape with parameter settings   given in Table 
 \ref{tab:problem}  and Table \ref{tab:truevalues} and create the respective problems synthetic ground-truth topography.

\begin{figure}[htbp!]
  \begin{center} 
  
  \includegraphics[width=160mm]{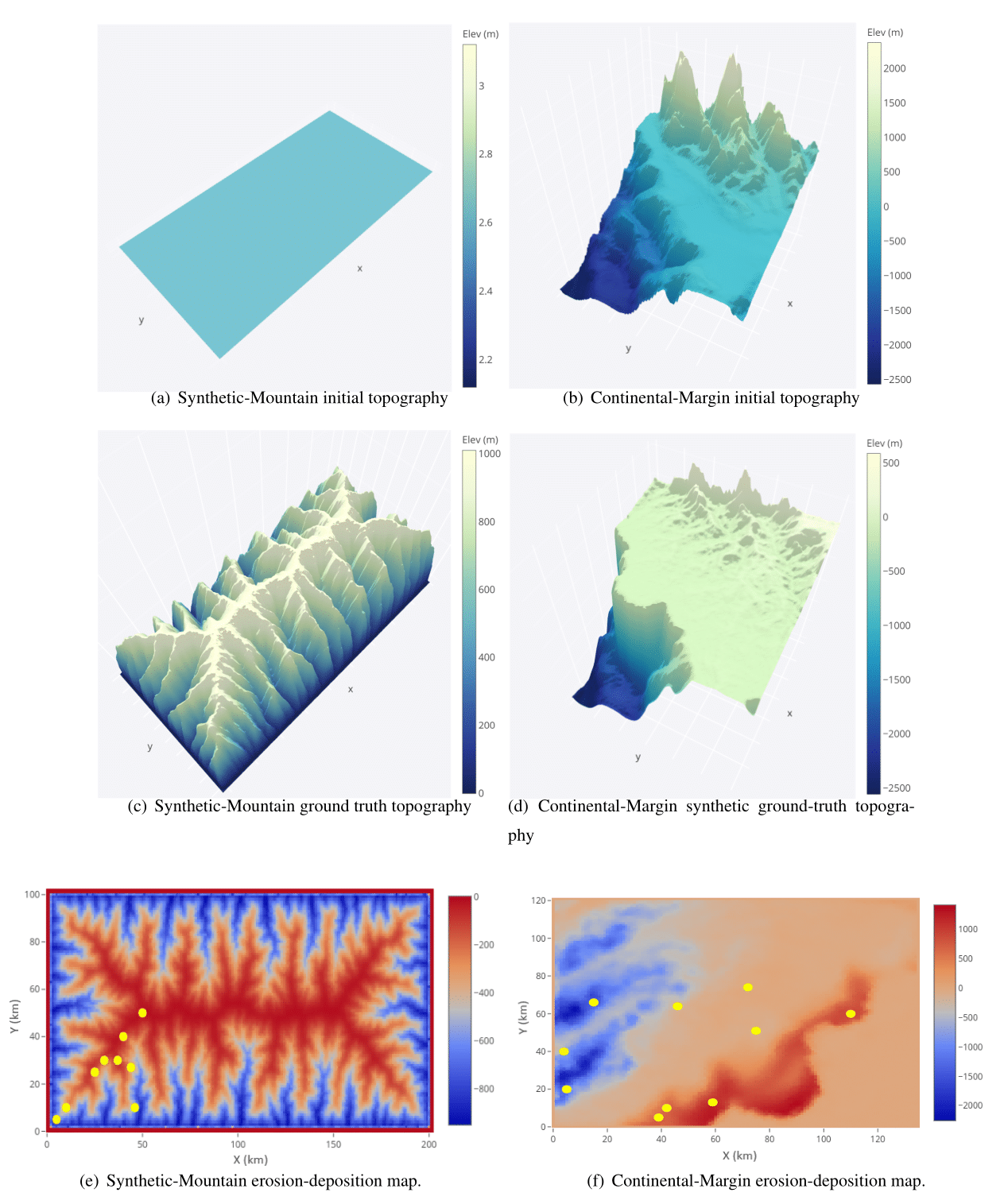}
    \caption{Synthetic-Mountain: Initial and eroded ground-truth topography after  a million years of evolution. Continental Margin  : 
Initial and eroded ground-truth topography and sediment after one million years.  The  erosion-deposition that forms sediment deposition after one million years is also shown.    Note that x-axis represents the latitude, y-axis represents the longitude and that aligns with Figure 1~Panel(a). The elevation in meters (m) is given by the z-axis which is further shown as a colour-bar. The Synthetic-Mountain problem does not align with actual landscape.  }
 \label{fig:craterdata}
  \end{center}
\end{figure}

\begin{figure}[htbp!]
  \begin{center} 
  
  \includegraphics[width=160mm]{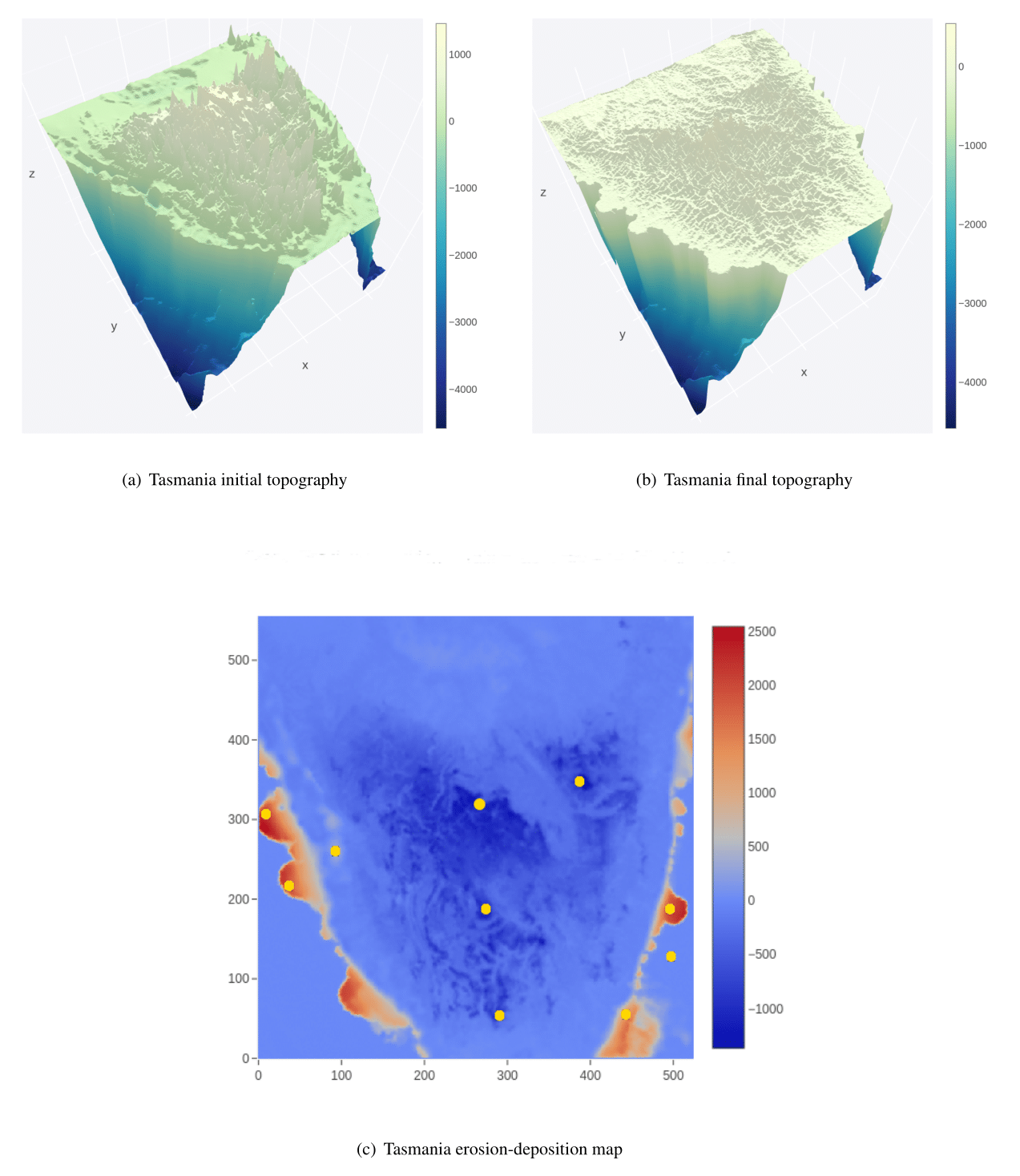}
    \caption{Tasmania: initial and eroded ground-truth topography along with erosion-deposition that shows sediment deposition after one million years evolution.    Note that x-axis represents the latitude, y-axis represents the longitude and that aligns with Figure 1~Panel(b) for the Tasmania problem. The elevation in meters (m) is given by the z-axis which is further shown as a colour-bar.}  
 \label{fig:tasmania_init_final}
  \end{center}
\end{figure}

   The initial and synthetic ground-truth topographies along with erosion/deposition  for these problems  appear 
in  Figure \ref{fig:craterdata} and \ref{fig:tasmania_init_final}, respectively. Note that the figure shows that the Synthetic-Mountain is flat in the beginning, then given a constant uplift rate along with weathering with constant precipitation rate creates the mountain topography. We use present-day topography as the initial topography in the Continental-Margin and Tasmania problems; whereas, a synthetic flat region for Synthetic-Mountain initial topography. The problems involve an erosion-deposition model history that is used to generate synthetic ground-truth data for the final model state that we then attempt to recover. Hence, the likelihood function given in the following subsection takes both the landscape topography and erosion-deposition ground-truth into account.    The 
Continental-Margin  and Tasmania cases  feature six free parameters (Table  
\ref{tab:truevalues}); whereas, the Synthetic-Mountain features 5 free parameters. Note that the marine diffusion coefficients are absent for the Synthetic-Mountain problem since the region does not cover or overlap with coastal and marine areas.  The main reason behind choosing the two benchmark problems is due to their nature, i.e. the Synthetic-Mountain problem features uplift rate, which is not present in the Continental-Margin problem. The Continental-Margin problem features other parameters such as the marine coefficients. The Tasmania problem features a much bigger region; hence, it takes more computational time for running a single model. The common feature in all three problems is that they model both the elevation and erosion/deposition topography. Furthermore, we draw the priors from a uniform distribution with a lower and upper limit given in Table  
\ref{tab:priors}.

\label{sec:syn_topo}

\begin{table*}[htbp!]
\small 
\centering
 \begin{tabular}{ l c c c c  c } 
 \hline 
Topography & Evo.(years)&  Length [km, pts] &  Width [km, pts]   & Res. factor & 
Run-time (s)  \\  
 \hline  
 Continental-Margin & 1 000 000  &  [136.0, 136] &  [123.0, 123]& 1 &  3.0\\ 
 Synthetic-Mountain  &  1 000 000 &  [ 202.0, 202 ] &  [102.0, 102 ]& 1  & 5.0 \\
 Tasmania & 1 000 000 &  [523.0,523]  &  [554.0,554] &  1  & 71.3  \\
 \hline
 \end{tabular}
 
\caption{In the given landscape evolution problems, the run-time 
represents approximately the duration for one model to run on a single CPU. The length and width is given in kilometers (km) that are represented by the specified number of points (pts) as defined by the resolution (Res.) factor.  }
 \label{tab:problem} 
\end{table*}

\begin{table*}[htbp!]
\small 
\centering
 \begin{tabular}{ l c c c c  c  c c } 
 \hline 
 \hline
Topography & Rainfall (m/a)&  Erod. &  n-value   & m-value & c-marine & 
c-surface & Uplift (mm/a)  \\  
 \hline

 Continental-Margin  & 1.5  & 5.0-e06  & 1.0  & 0.5 &  0.5  & 0.8 & - \\  
Synthetic-Mountain  & 1.5  & 5.0-e06  & 1.0  & 0.5 &  -  & - & 1.0 \\  
 Tasmania & 1.5  & 5.0-e06  & 1.0  & 0.5 &  0.5  & 0.8 & -  \\  
 \hline

 \end{tabular}
 
\caption{True values of parameters  }
 \label{tab:truevalues} 
\end{table*}

\begin{table*}[htbp!]
\small 
\centering
 \begin{tabular}{ l c c c c  c  c c } 
 \hline 
 \hline
Topography & Rainfall (m/a)&  Erod. &  n-value   & m-value & c-marine & 
c-surface & uplift  \\  
 \hline   
 Continental-Margin & [0,3.0 ]  & [3.0-e06, 7.0-e06] & [0, 2.0]  & [0, 2.0] & [0.3, 0.7]  & 
[0.6, 1.0] & - \\ 
Synthetic-Mountain & [0,3.0 ]  & [3.0-e06, 7.0-e06] & [0, 2.0] & [0, 2.0]  & -  & - & 
[0.1, 1.7] \\ 
Tasmania & [0,3.0 ]  & [3.0-e06, 7.0-e06] & [0, 2.0]  & [0, 2.0] & [0.3, 0.7]  & 
[0.6, 1.0] & - \\

 \hline
 \end{tabular}
\caption{Prior distribution range of model parameters }
 \label{tab:priors} 
\end{table*}

\subsection {Bayeslands likelihood function }

 The Bayeslands likelihood function evaluates  Badlands topography simulation along with the successive erosion-deposition which denotes the sediment thickness evolution through time.   More specifically, the likelihood function evaluates the effect  of the proposals by taking into account the difference between the final simulated Badlands topography and the ground-truth topography.   The likelihood function also considers the difference between the simulated and ground-truth sediment thickness at selected time intervals, which has been adapted from previous work  \citep{chandra2018PT-Bayes_} and  given as follows.   The initial topography is  denoted by $\boldsymbol D_0$ with $\boldsymbol D_0=(D_{0,s_1}\ldots,D_{0,s_n})$, where $s_i$ corresponds to site $s_i$, with the coordinates given by the latitude  $u_i$  and longitude  $v_i$. 
 
 We assume an inverse gamma (IG) prior  $\tau^2\sim IG(\nu/2,2/\nu)$ and integrate it  so that  the likelihood for the topography at time $t=T$ is  
 
\begin{equation}
 L_{l}(\boldsymbol\theta)\propto \prod_{i=1}^n \left(1+\frac{(D_{s_i,T}-f_{s_i,T}(\boldsymbol\theta))^2}{\nu}\right)^{-\frac{\nu+1}{2}}
\end{equation}
where $\nu$ is the number of observations and  the subscript $l$, in $L_l(\boldsymbol\theta)$  denotes that it is the landscape likelihood to distinguish it from a sediment likelihood.

Although Badlands produces successive time-dependent topographies, only the final topography $\mathbf D_T$ is used for the calculation of the elevation likelihood since little ground-truth information is available for the detailed evolution of surface topography. In contrast,    the time-dependence of sedimentation  can be used to ground-truth the time-dependent evolution of surface process models that include sediment transportation and deposition. The  sediment erosion/deposition
values at time  ($\mathbf z_t$) are simulated (predicted) by the Badlands   model  given  set of parameters,
$\boldsymbol \theta$ plus some Gaussian noise
\begin{equation}
z_{s_j,t} =g_{s_j,t}(\boldsymbol \theta)+\eta_{s_j,t}\;\mbox{with}\; \eta_{s_j,t}\sim 
(0,\chi^2)
\label{eqn:sed_sim}\end{equation}
The sediment likelihood  $L_s(\boldsymbol\theta)$, after integrating out $\chi^2$  becomes
\begin{equation}
 L_{s}(\boldsymbol \theta)\propto \prod_{t=1}^T\prod_{j=1}^J \left(1+\frac{(z_{s_j,t}-g_{s_j,t}(\boldsymbol \theta))^2}{\nu}\right)^{-\frac{\nu+1}{2}}
\end{equation}

 The combined likelihood  takes  both elevation and sediment/deposition into account 
 \begin{equation}
 L(\boldsymbol \theta)=L_s(\boldsymbol \theta)\times L_l(\boldsymbol \theta).
 \end{equation} 
 
  Note that although we used the log-likelihood version in our actual implementation, we refer to it as the likelihood throughout the paper.

 \begin{figure}[htbp!]
  \begin{center} 
  
  \includegraphics[width=140mm]{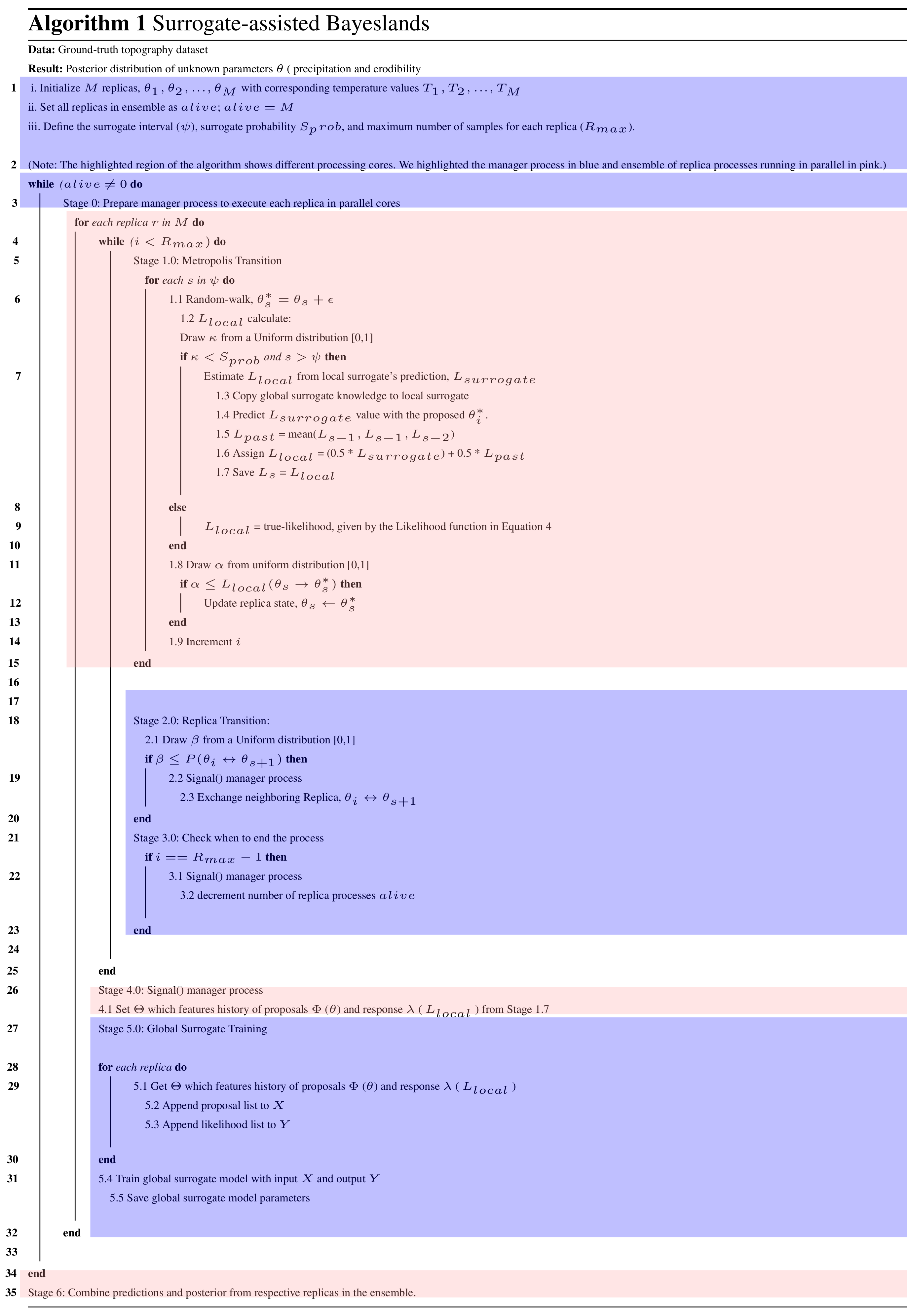}
     
 \label{alg:ptfnn}
  \end{center}
\end{figure}

 \subsection{Surrogate-assisted   Bayeslands}

The surrogate model learns from the relationship between the set of input parameters and the response given by the true (Badlands) model.  The input is the set of proposals by the respective replica  samplers in the parallel tempering MCMC sampling algorithm. We refer to the likelihood estimation by the surrogate model as the \textit{pseudo-likelihood}. 

We need to take into account the cost of inter-process communication in parallel computing environment   to avoid computational overhead. As given in our previous implementation \citep{chandra2018PT-Bayes_}, the  \textit{swap interval}  refers to the number of iterations after which each replica pauses and can undergo a replica transition. After the swap proposal is accepted or rejected, the respective replica sampling is  resumed while undergoing Metropolis transition in between the swap intervals. We incorporate the surrogate-assisted estimation into the multi-core parallel tempering algorithm. Our previous work \citep{chandra2020SAPT} used a \textit{surrogate interval} that determines the frequency of training by collecting the history of past samples with their likelihood from the respective replicas. We need a  swap interval of several samples   when dealing with small scale models that take a few seconds to run; however for large models, we recommend to have a swap interval of 1.

Taking into account that the  true model is represented as $y = f(x)$, the surrogate model provides an approximation in the form $\hat{y} = \hat{f}(x) $; such that $y = \hat{y} + e$, where  $e$ represents the difference or error. The task of the surrogate model is to provide an estimate for the pseudo-likelihood by training from the history of proposals which is given by the set of input $\bf{x}_{r,s}$ and likelihood $y_s$ where $s$ represents the sample and $r$ represents the replica. Hence, we create the training dataset $\Phi $ for the surrogate by fusion of $\bf{x}_{r,s}$  across all the replica for a given surrogate interval $\psi$,  which can be formulated as follows 

\begin{eqnarray}
\Phi&=&(\bf{x}_{1,s},\ldots,\bf{x}_{1,s+\psi}, \ldots, \bf{x}_{M,s},\ldots,\bf{x}_{M,s+\psi})\nonumber\\
\lambda &=&(y_{1,s},\ldots,y_{1,s+\psi}, \ldots, y_{M,s},\ldots,y_{M,s+\psi})
\label{data}
\end{eqnarray}
 where, $\bf{x}_{r,s}  $ represents the set of parameters proposed  at sample $s$, $y_{r,s} = \log\left(p({\bf y }|\boldsymbol{\bf{x}_{r,s}})\right)$ is the  likelihood which is dependent on data and the Badlands model, and $M$ is the total number of replicas.
$\Theta$ denotes the  training surrogate dataset  which features input $\Phi$ and response $ \lambda$ at the end of every surrogate interval denoted by $s+\psi$.  Therefore, we give the pseudo likelihood as $\hat{y} = \hat{f}(\Theta) $, where $\hat{f}$ is the prediction from the surrogate model.
 The likelihood in training data is altered, with respect of the temperature,  since it has been changed by taking ${L_{local}}/{T_r}$ for given replica $r$. We undo this change by multiplying the likelihood by the respective replica temperature level taken from the geometric temperature ladder. 
  
\begin{figure*}[htb]
  \begin{center} 
  \includegraphics[width=150mm]{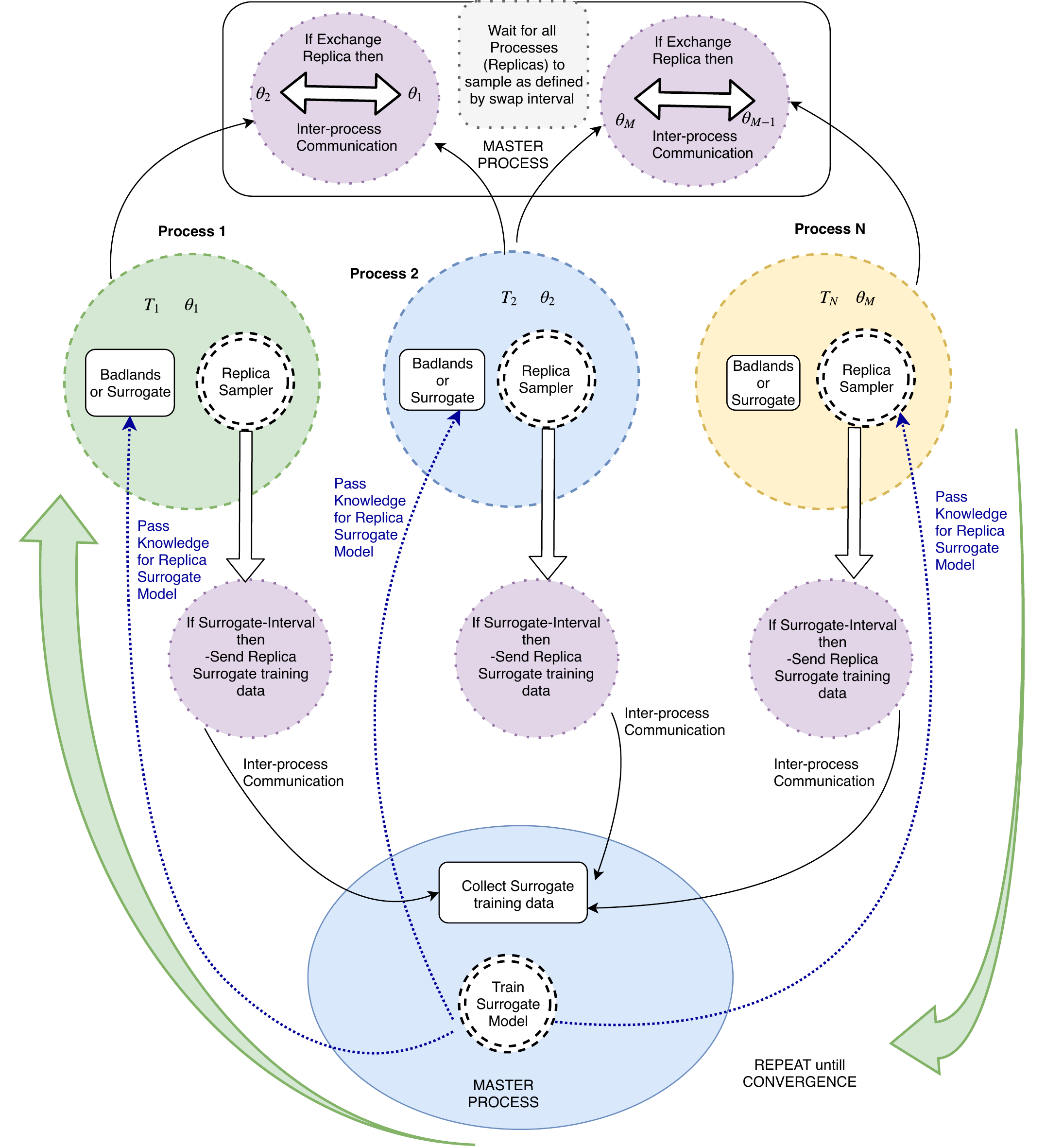}
    \caption{ Surrogate-assisted Bayeslands using the parallel tempering MCMC framework.  We carry out the training in the master (manager) process which features the global surrogate model. The replica processes provide the surrogate training dataset to the master process using inter-process communication.   We employ a neural network model  for the surrogate model. After training, we transfer the knowledge (neural network weights)  to each of the replicas to enable  estimation of pseudo-likelihood. Refer to Algorithm 1 for further details.} 
 \label{fig:surr_framework}
  \end{center}
\end{figure*}

We present surrogate-assisted Bayeslands in Algorithm \ref{alg:ptfnn} that features parallel processing of the ensemble of replicas. The highlighted region in colour pink of the Algorithm 1 shows different processing cores running in parallel,  shown in Figure \ref{fig:surr_framework} where the manager process is highlighted. Due to multiple parallel processing replicas,  it is not straightforward to implement when to terminate sampling. Hence, the termination condition waits for all the replica processes to end as it  monitors the number of active or \textit{alive replica processes}  in the manager process. We begin by setting the number of alive replicas in the ensemble   ($alive = M$) and then the replicas that sample $\theta_n$    are assigned values using a uniform distribution  $[-\alpha,\alpha]$; where $\alpha$ defines the range of the respective parameters.       We then assign the user-defined  parameters   which include the number of replica samples   $R_{max}$, swap-interval   $R_{swap}$,  surrogate interval, $\psi$, and   surrogate probability  $S_{prob}$ which determines the frequency of employing the surrogate model for estimating the pseudo-likelihood.

The samples that cover the first surrogate interval makes up the initial surrogate training data $\Theta$, which features all the replicas.  We then train the surrogate to estimate the pseudo-likelihood when required according to the surrogate probability.  Figure \ref{fig:surr_framework} shows how the manager processing unit controls the respective replicas, which samples for the given surrogate interval. Then, the algorithm calculates the replica transition probability for the possibility of swapping the neighbouring replicas.     The information flows from replica process to manager process using \textit{signal()} via inter-process communication given by the replica process as shown in Stage 2.2,  3.1 and 4.0 of Algorithm 1, and further shown in Figure \ref{fig:surr_framework}.

 To enable better estimation for the pseudo-likelihood, we retrain the surrogate model for remaining surrogate interval blocks until the maximum time ($R_{max}$). We train the surrogate model   only in the manager process and  the algorithm passes the surrogate model copy with the trained parameters to the ensemble of replica processes for predicting or estimating the pseudo-likelihood.  The samples associated with the true-likelihood only becomes part of the surrogate training dataset.  In   Stage 1.4 of Algorithm 1, the  pseudo-likelihood ($L_{surrogate}$) provides an estimation with given proposal $\theta_s^*$. Stage 1.5 calculates the likelihood moving average of past three likelihood values,  $L_{past}$ = mean($L_{s-1}, L_{s-1}, L_{s-2}$). In Stage 1.6, we combine the moving average likelihood  with the pseudo-likelihood to give a prediction that considers the present replica proposal and taking into account the past, $L_{local}$ =  (0.5 * $L_{surrogate}$) + 0.5 * $L_{past}$. The surrogate training can consume a significant portion of time which is dependent on the size of the problem in terms of the number of parameters and also the type of surrogate model used, along with the training algorithm.  We   evaluate the  trade-off between   quality of estimation by pseudo-likelihood and overall cost of computation for the true likelihood function for different types of problems.

We validate the quality of estimation from the surrogate model by the root mean squared error (RMSE) which considers the difference between the true likelihood and the pseudo-likelihood. This can be seen as a regression problem with multi-input (parameters) and a single output (likelihood).  Hence, we  report the  surrogate prediction quality   by

$$RMSE_{sur} = \sqrt{\frac{1}{N} \sum_{i=1}^{N} (y_i - \hat{y_i})^2}$$

where $y_i$ and $\hat{y_i}$ are the true likelihood and the pseudo-likelihood values, respectively. $N$ is the number of cases the surrogate is used during sampling.

 We further note that the framework uses parallel tempering MCMC in the first stage of sampling and then transforms into the second stage where the temperature ladder is changed such that $T_i = 1$, for all replicas, $i = 1, 2, ..., M$.  This strategy enables exploration is the first stage and exploitation in the second stage.  We combine the respective replica posterior distributions  once the termination condition is met and show their mean and standard deviation of the prediction in the results.

We evaluate the prediction performance   by comparing the  predicted/simulated Badlands  landscape with the ground-truth data using  the root-mean squared error (RMSE). We compute the RMSE  for the  elevation (elev) and sediment erosion/deposition (sed)  at each iteration of the sampling scheme using

\begin{eqnarray*}
 \mbox{RMSE}_{elev} & = &\sqrt{\frac{1}{n\times m}\sum_{i=1}^n\sum_{j=1}^n 
\left(g(\hat{\theta}_{T,i,j}) -g_{T,i,j}(\theta)\right)^2}\\
 \mbox{RMSE}_{sed} & = & \sqrt{\frac{1}{n_t \times 
v}\sum_{t=1}^{n_t}\sum_{j=1}^m \left(f(\hat{\theta}_{t,j}) 
-f(\theta_{t,j})\right)^2}
  \end{eqnarray*}

 where, $\hat{\theta}$ is an estimated value of $\theta$, 
 and $\theta$ is the true value  representing the synthetic ground-truth.  $f(.)$ and $g(.)$ represent 
the outputs of the  Badlands model  while $m$ and $n$ 
represent the size of the selected topography.  $v$ is the  number of 
selected points from   sediment erosion/deposition over the selected time frame, 
$n_t$. 
 \subsection{Surrogate model}
 
 To choose a particular surrogate model, we need to consider the computational resources for training the model during the sampling process. The literature review showed that Gaussian process models, neural networks, and radial basis functions \citep{broomhead1988radial} are popular choices for surrogate models.  We note that Badlands LEM features about a dozen of free parameters in one of the simplest cases, this increases when taking into account spatial and temporal dependencies. For instance, the precipitation rate for a million years can be represented by a single parameter or by 10 different parameters that capture every 100,000 years for 10 different regions, which can account for 1,000 parameters instead of 1. Considering hundreds or thousands of unknown Badlands model parameters,  the surrogate model needs to be efficiently trained without taking lots of computational resources. The flexibility of the model to have incremental training is also needed and hence, we rule out Gaussian process models since they have  limitations in training when  the size of the dataset increases to a certain level \citep{rasmussen2004gaussian}. Therefore, we use neural networks as the choice of the surrogate model and the training data and neural network model is formulated as follows.

We denote the surrogate model   training data   by $\Phi$ and $\lambda$ which is shown in Equation \eqref{data}; where $\Phi$ is the input and $\lambda$ is the desired output of the model. The prediction of the model is denoted by $\hat{\lambda}$. 
 We use a feedforward neural network as the surrogate model. Given input  $\bf x_t$,   $f({\bf x}_t)$ is computed by the  feedforward neural network with one hidden layer 
defined by the function

\begin{equation}
f({\bf x}_t)   =   g \bigg(  \delta_o + 
\sum_{h=1}^{H} v_{j} g \bigg(  \delta_h + \sum_{d=1}^{I} w_{dh} 
\bf x_{t} \bigg)\bigg) 
\label{expected_y}
\end{equation}

where $\delta_o$ and $\delta_h$  are the bias weights for the output $o$ and 
hidden $h$ layer, respectively.  $v_j$ is the weight which maps the hidden 
layer $h$ to the output layer. $w_{dh}$ is the weight which maps $\bf x_{t}$ to 
the hidden layer $h$ and $g(.)$ is the activation function 
 for the hidden and output layer units.  We use ReLU (rectified linear unitary function) as the activation function. The learning or optimisation task then is to iteratively update the weights and biases to minimise the cross-entropy loss $J(\bf{W,b})$. This can be  done using gradient update of weights using  Adam (adaptive moment estimation) learning algorithm \citep{kingma2014adam} and  stochastic gradient descent \citep{bottou1991stochastic,bottou2010large}. We experimentally evaluate them for training feedforward network for the surrogate model in the next section. 

  \subsection{Proposal distribution}
  
  Bayeslands features  random-walk (RW) and adaptive random-walk (ARW) proposal distributions which will be evaluated further for surrogate-assisted Bayeslands in our experiments. In our previous work \citep{Chandra2018_Bayeslands_}, AWR showed better convergence properties when compared to RW proposal distribution. The RW proposal distribution features   $\Sigma$  as the diagonal matrix, so that $\Sigma=\mbox{diag}(\sigma^2_1,\ldots, \sigma^2_P)$; where $\sigma_j$ is the step size of the $j^{th}$ element of the parameter vector $\boldsymbol \theta$.  The step-size for   $\theta_j$ is a combination of a fixed step size $\phi$ which is  common to all parameters,
 multiplied by the range of possible values for parameter $\theta_j$, hence $\sigma_j = (a_j - b_j) \times \phi $; where, $a_j$ and $b_j$ represent the maximum and minimum limits of the prior for $\theta_j$   given in Table \ref{tab:truevalues}. In our experiments, the RW proposal distribution employs fixed   step-size,   $\phi=0.05$,
 
 The ARW proposal distribution features adaptation of the diagonal matrix  $\Sigma$ at  every $K$ interval  of within-replica sampling.  It allows for the dependency between elements of $\boldsymbol \theta$ and adapts  during sampling \citep{haario2001adaptive}.  We adapt the elements of $\Sigma$  for the posterior distribution using the sample covariance of the current chain history  $\Sigma = \mbox{cov}(\{\boldsymbol \theta^{[0]}, \ldots, \boldsymbol  \theta^{[i-1]}\}) + \mbox{diag}(\lambda_1^2,\ldots,\lambda_P^2)$; where $\boldsymbol \theta^{[i]}$ is the $i^{th}$ iterate of $\boldsymbol \theta$ in the chain and $\lambda_j$ is the minimum allowed step sizes for each parameter $\theta_j$.

\subsection{Design of Experiments}

We demonstrate effectiveness of  surrogate-assisted parallel tempering (SAPT-Bayeslands) framework for selected Badlands LEMs taken from our  previous study \citep{chandra2018PT-Bayes_}. 

  We first investigate the effects of different surrogate training procedures and parameter evaluation for SAPT-Bayeslands using smaller synthetic problems. Afterwards, we apply the methodology to a larger landscape evolution problem which is Tasmania, Australia. We design the experiments as follows. 
 
\begin{enumerate}

\item  We generate a dataset for training and testing the surrogate for the  Synthetic-Mountain and Continental-Margin landscape evolution problems. We use the neural network model for the surrogate and evaluate different training techniques.

\item  We evaluate if the transfer of knowledge from previous surrogate interval is better than no transfer of knowledge for Synthetic-Mountain and Continental-Margin problems. Note this is done only with the data generated from the previous step.

\item  We provide convergence diagnosis for the RW and ARW proposal distributions in PT-Bayeslands and SAPT-Bayeslands. 

\item We integrate the surrogate model into  Bayeslands    and evaluate the effectiveness of the surrogate in terms of estimation  of the likelihood and computational time. Due to the computational requirements, we only consider Continental-Margin  problem.

\item  We then apply SAPT-Bayeslands   to  all the  given problems and compare with PT-Bayeslands.

\end{enumerate}

  We   use \textit{Keras} neural networks library \citep{chollet2015keras} for implementation of the surrogate.   We provide the open-source  software package that implements Algorithm 1 along with benchmark problems and experimental results \footnote{Surrogate-assisted Bayeslands:  
\url{https://github.com/badlands-model/surrogateBayeslands} }. 

We use a  geometric temperature ladder with a maximum temperature of $T_{max} = 2 $  for determining the temperature level for each of the replicas.   In trial experiments, the selection of these parameters depended on the performance in terms of the number of accepted samples and prediction accuracy of elevation and sediment/deposition.   We use  replica-exchange or swap interval value,
$R_{swap} = 3$ samples that determine when to check whether to swap with the neighbouring replica.  In previous work \citep{chandra2018PT-Bayes_},  we observed that increasing the number of replicas up to a certain point does not necessarily mean that we get better performance in terms of the computational time or prediction accuracy. In this work, we limit the number of replicas as $R_{num} =8$ for all experiments  with maximum   of 5,000 samples.

We use a 50 \% burn-in which discards the portion of samples in the parallel tempering MCMC stage as done in our previous work \citep{Chandra2018_Bayeslands_}.

\section {Results}
\subsection{Surrogate accuracy}
 
  To implement the surrogate model, we need to evaluate the training algorithm such as Adam and stochastic gradient descent (SGD). Furthermore, we also evaluate specific parameters such as the size of the surrogate interval (batch-ratio), the neural network topology for the surrogate and the effectiveness of either training from scratch or to utilise previous knowledge for surrogate training (transfer and train).  We create a training dataset from the cases where the true likelihood was used, which compromises the history of the set of parameters proposed with the corresponding likelihood.  This is done for standalone evaluation of the surrogate model, which further ensures that the experiments are reproducible since different experimental runs create different dataset depending on the exploration during sampling.   We then evaluate the neural network model designated for the surrogate using two major training algorithms which featured the Adam optimiser and stochastic gradient descent. The parameters that define the neural network surrogate model used for the experiments are given in Table \ref{tab:param}. Note that the train size in  Table \ref{tab:param} refers to the maximum size of the data set. The training is done in batches where the batch ratio determines the training data set size, as shown in Table \ref{tab:accuracy}.  

\begin{table}[!h]
\centering
\small
 \caption{Neural network architecture for the different problems}
\label{tab:param}
\begin{tabular}{ l l  l l l}
    \hline
 	\hline
	Dataset         & Input & Output &    Train size & Test size\\
	\hline
	\hline
	Continental-Margin		& 6     & 1        & 8073 & 879\\ 
	Synthetic-Mountain 		& 5     & 1       & 8073 & 879\\
	\hline
\end{tabular}
\end{table}

\begin{table*}
\centering
\caption{Evaluation of surrogate training accuracy}
\label{tab:accuracy}
\begin{tabular}{ l  l  l  l  l  l l  l  l  l }
\hline \hline
	Dataset & Batch-ratio &  \multicolumn{4}{c}{Transfer and train }  &  \multicolumn{4}{c}{Train from scratch}  \\ 
	\hline
	 &  & \multicolumn{2}{c}{SGD} & \multicolumn{2}{c}{Adam} & \multicolumn{2}{c}{SGD} & \multicolumn{2}{c}{Adam} \\ \hline
	 
	 &  & MSE & Time(s) & MSE & Time(s) & MSE & Time(s) & MSE & Time(s) \\ \hline \hline
Continental-Margin & 0.1 & 0.0198 & 19.40 & 0.0209 & 31.23 & 0.0199 & 88.17 & 0.0206 & 122.41\\ 
	 & 0.2 & 0.0197 & 26.95 & 0.0211 & 56.84& 0.0197 & 67.74 & 0.0199 & 100.49 \\ 
	 & 0.3 & 0.0199 & 25.53 & 0.0212 & 61.41 & 0.0197 & 70.71& 0.0205 & 268.16\\ 
	 & 0.4 & 0.0195 & 70.42& 0.0193 & 48.28 & 0.0194 & 46.07 & 0.0188 & 140.90 \\ \hline   
	Synthetic-Mountain & 0.1 & 0.0161 & 40.38& 0.0097 & 54.45 & 0.0161 & 282.0 & 0.0081 & 347.94\\ 
	 & 0.2 & 0.0134 & 52.87& 0.007 & 70.65 & 0.0139 & 185.025 & 0.007 & 857.38 \\ 
	 & 0.3 & 0.0129 & 65.105 & 0.0088 & 73.035 & 0.0123 & 179.36 & 0.0088 & 543.019 \\ 
	 & 0.4 & 0.0164 & 50.14 & 0.0048 & 87.67 & 0.0066 & 149.26 & 0.0038 & 653.85 \\ \hline \hline 
\end{tabular}
\end{table*}

  Table \ref{tab:accuracy} presents the results for the experiments that took account of the training data collected during sampling for two benchmark problems (Continental-Margin and  Synthetic-Mountain). Note that, we report the mean value of the mean-squared-error (MSE) for the given batch ratio from ten experiments. The batch ratio is taken, in relation to the maximum number of samples across the chains ($R_{max}/R_{num}$). We normalise the likelihood values (outcomes) in the dataset between [0,1].    Although in most cases, the accuracy of the neural network is slightly better when training from scratch with combined data; however,  there is a considerable trade-off with the time required to train the network. The results show that the transfer and train methodology, in general, requires much lower computational time when compared to training from scratch by combined data.
Moreover, in comparison to SGD and Adam training algorithms, we observe that   SGD achieves slightly better accuracy than Adam for Continental-Margin problem. However, Adam, having an adaptive learning rate, outperforms SGD in terms of the time required to train the network. Thus, we  can summarise that transfer and train method is better since it saves significant computation time with a minor trade-off with accuracy.

\subsection{Convergence diagnosis }

 The Gelman-Rubin diagnostic  \citep{gelman1992inference} is one of the popular methods used for evaluating convergence by analyzing the behaviour of  multiple Markov chains. The assessment is done  by comparing the estimated between-chains and within-chain variances for each  parameter, where large differences between the variances indicate non-convergence.  The diagnosis reports  the potential scale reduction factor (PSRF) which  gives the ratio of the current variance in the posterior variance for each parameter compared to that being sampled and the values for the PSRF near 1 indicates convergence.    We analyse five experiments for each case  using  different initial values for 5,000 samples for each   problem configuration.

Table \ref{tab:convergence} presents the convergence diagnosis using the PSRF score for RW and ARW proposal distributions for PT-Bayeslands and SAPT-Bayeslands.  We notice that ARW has lower PSRF score (mean) when compared to RW proposal distribution which indicates better convergence. We also notice that the ARW SAPT-Bayeslands maintains convergence with PSRF score close to AWR PT-Bayeslands when compared to rest of the  configurations. This suggests that although we use surrogates, convergence can be maintained up to a certain level, which is better than RW PT-Bayeslands.

\begin{table*}[htbp!]
\small
\centering
\caption{Convergence diagnosis (PSRF score)  for Continental Margin problem }
\begin{tabular}{ l l l l l l l l l  }
\hline \hline 
	 Proposal & method & Precip. & Erod. & m-value  & n-value & c-marine & c-surface & Mean R-Score   \\ 
	 
	 \hline \hline 
	 
RW &   PT-Bayeslands  & 1.50 & 1.6 & 1.14 &  4.82 & 2.62 & 1.56 & 2.21 \\

ARW &  PT-Bayeslands  & 1.26 &  1.55 &  1.26 &  1.63 & 1.38 &  1.13 &  1.37 \\
\hline

RW &   SAPT-Bayeslands  &  4.06 &  1.70 & 6.57 &  1.51 &  1.46 &  1.49 & 2.80 \\
ARW &      SAPT-Bayeslands &  1.33 &  2.88 &  1.22 & 2.46 &  1.03 &  1.30 &  1.70 \\

     \hline \hline
\end{tabular}

\label{tab:convergence}
\end{table*}

\subsection{Surrogate-assisted   Bayeslands } 
  
We investigate the effect of the surrogate probability ($S_{prob}$)   and surrogate interval ( $\psi$) on the prediction accuracy ($RMSE_{elev}$ and $RMSE_{sed}$) and computational time.   Note that we report the prediction accuracy  mean and standard deviation (mean and std)  of accepted samples   over the sampling time after removing the burn-out period.  We report the computational time in seconds (s). Table \ref{tab:surrogatecmargin}  presents the performance of the respective methods (PT-Bayeslands and SAPT-Bayeslands) with respective parameter settings for the Continental-Margin problem.  In SAPT-Bayeslands, we observe that there not a major  difference in the accuracy of elevation or erosion/deposition given different values of $S_{prob}$. Nevertheless, there is a significant difference in terms of the computational time where  higher  values of  $S_{prob}$  saves computational time.  Furthermore, we notice that there is not a significant difference in the prediction accuracy  given different values of   $\psi$ which suggests that the selected values are sufficient.

  We select a suitable combination of the set of parameters evaluated in the previous experiment ($S_{prob}=0.6$ and $\psi=0.05$) and apply to rest of the problems.  Table \ref{tab:restasmania} gives a comparison of performance for Continental-Margin and Synthetic-Mountain problem, along with the   Tasmania  which is a bigger and computationally expensive problem. We   notice that the performance of SAPT-Bayeslands is similar to PT-Bayeslands while a significant portion of computational time is saved.

\begin{figure*}[htbp!]
  \begin{center} 
  \includegraphics[width=160mm]{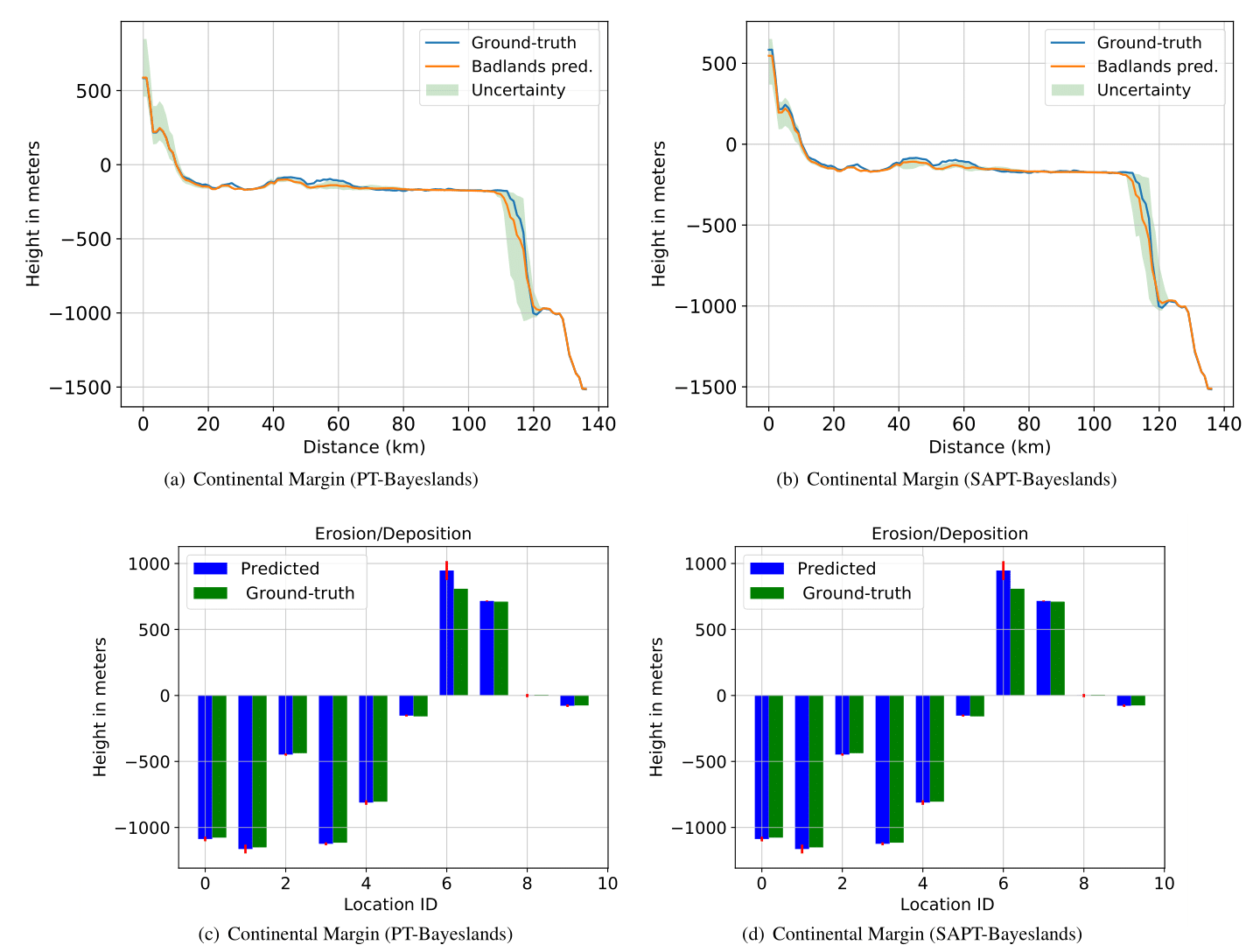}
    \caption{ Topography cross-section   and  erosion-deposition prediction    for 10 chosen points  (selected coordinates denoted by location identifier (ID) number)   for  Continental-Margin problem from results summarized in Table 8. }
 \label{fig:crosssectionxy_cm}
  \end{center}
\end{figure*}

\begin{figure*}[htbp!]
  \begin{center} 
  
  \includegraphics[width=160mm]{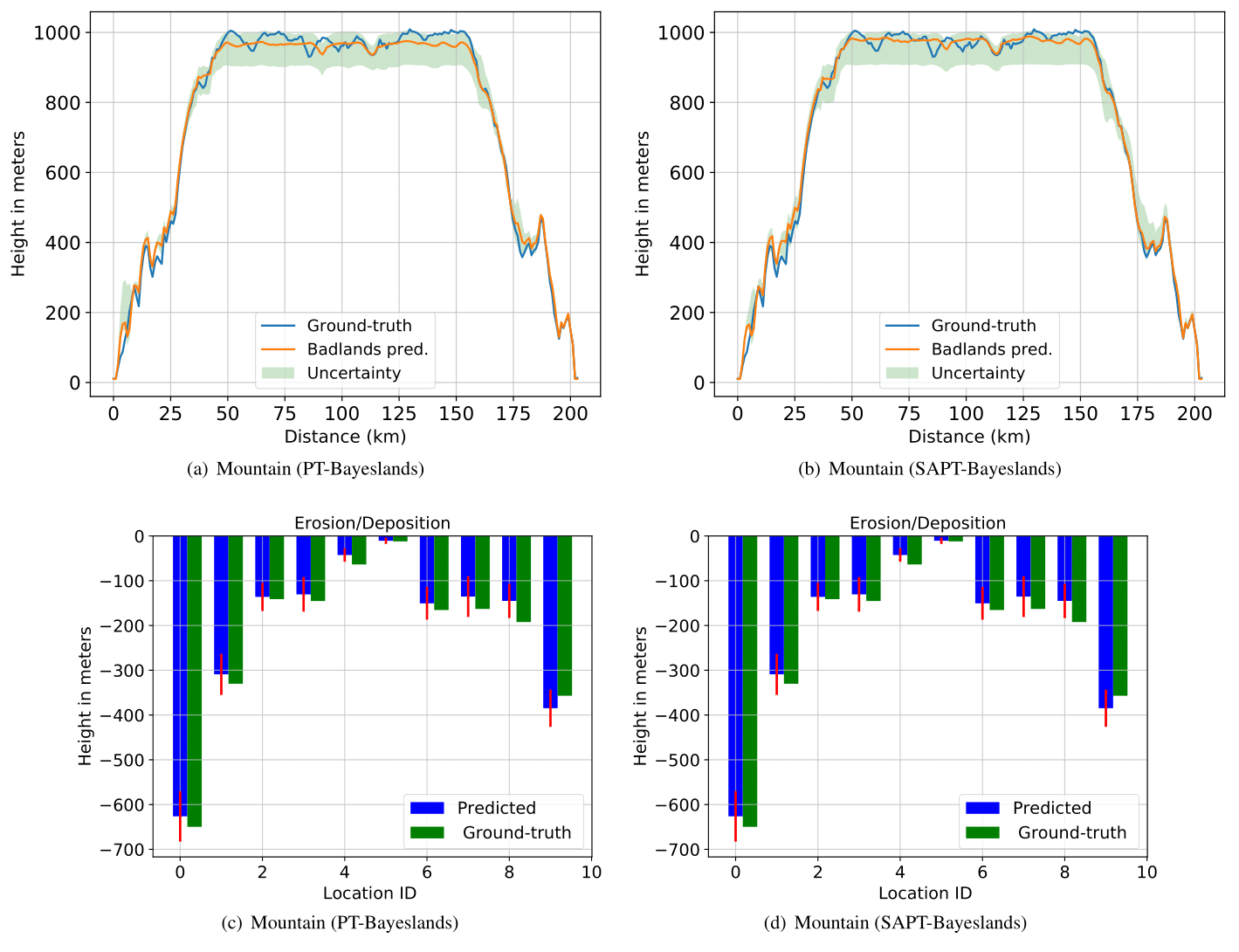}
    \caption{ Topography cross-section   and  erosion-deposition prediction    for 10 chosen points  (selected coordinates denoted by location identifier (ID) number)   for  Synthetic-Mountain problem from results summarized in Table 8.}
 \label{fig:crosssectionxy_mn}
  \end{center}
\end{figure*}

\begin{figure*}[htbp!]
  \begin{center} 
  \includegraphics[width=160mm]{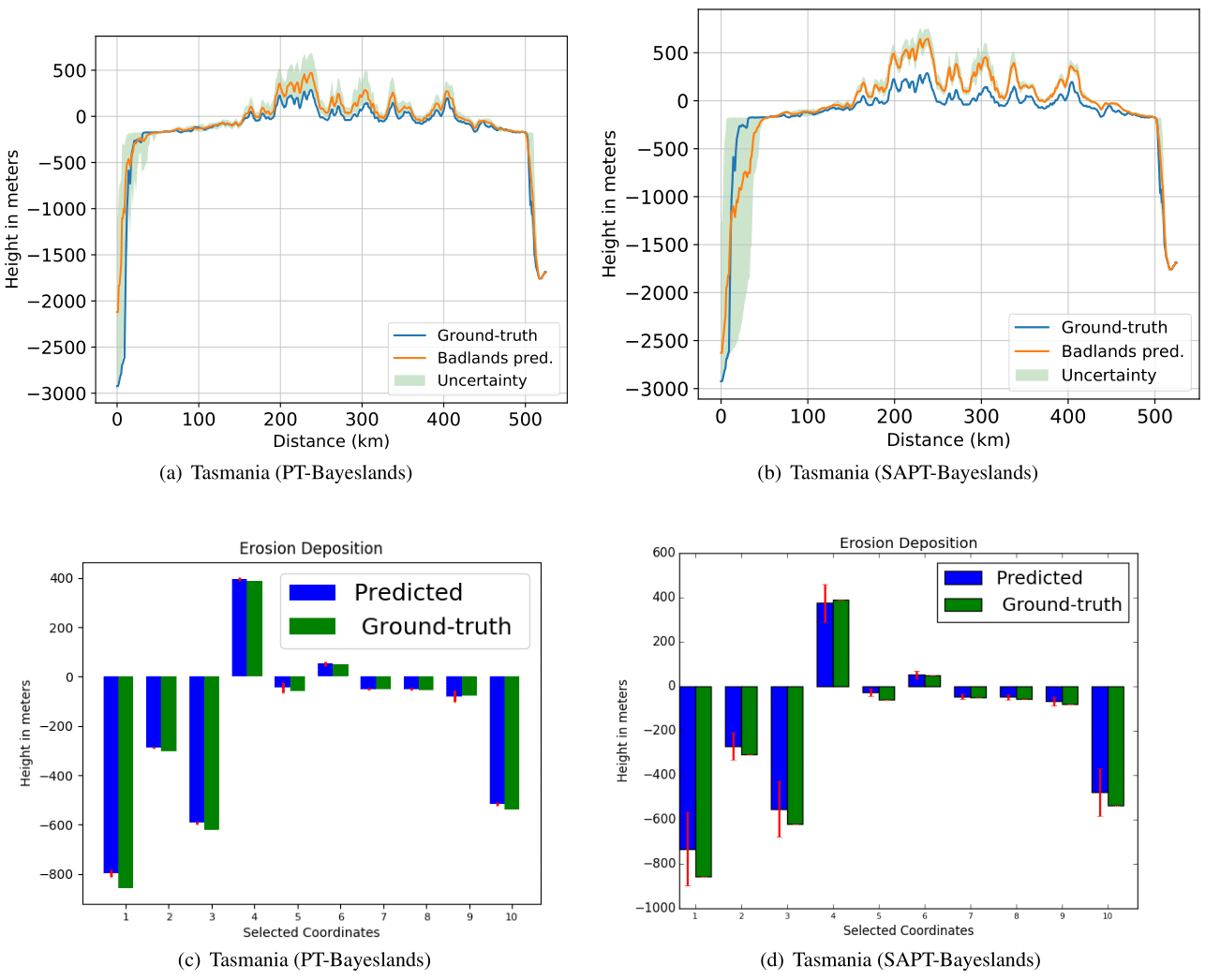}
    \caption{  Topography cross-section   and  erosion-deposition prediction    for 10 chosen points (selected coordinates denoted by location identifier (ID) number)  for Tasmania problem from results summarized in Table 8. }
 \label{fig:crossxy_tasmania}
  \end{center}
\end{figure*}

Figures \ref{fig:crosssectionxy_cm}, \ref{fig:crosssectionxy_mn} and \ref{fig:crossxy_tasmania} provides a   visualization in the elevation prediction accuracy when compared to actual ground-truth between the given methods from results given in Table \ref{tab:restasmania}.  We also provide the prediction accuracy of erosion/deposition for 10 chosen points taken at selected locations. Although both methods provide erosion/deposition prediction for 4 successive time intervals, we only show the final time interval.   In both the Continental Margin and Synthetic Mountain problems, we notice that although the prediction accuracy of PT-Bayeslands is very similar  to SAPT-Bayeslands and the Badlands prediction of the topography is close to ground-truth, within the credible interval. This indicates that the use of surrogates has been beneficial where not major loss in accuracy in prediction is given.  In the case of the Tasmania problem, there is a  loss in badlands prediction accuracy which could be due to the size of the problem. Nevertheless, this loss is not that clear from results in Table \ref{tab:restasmania}. This could be that the topography prediction is mostly inconsistent at the cross-section where  it features mountainous regions. 

\begin{figure*}[htbp!]
  \begin{center} 
  \includegraphics[width=100mm]{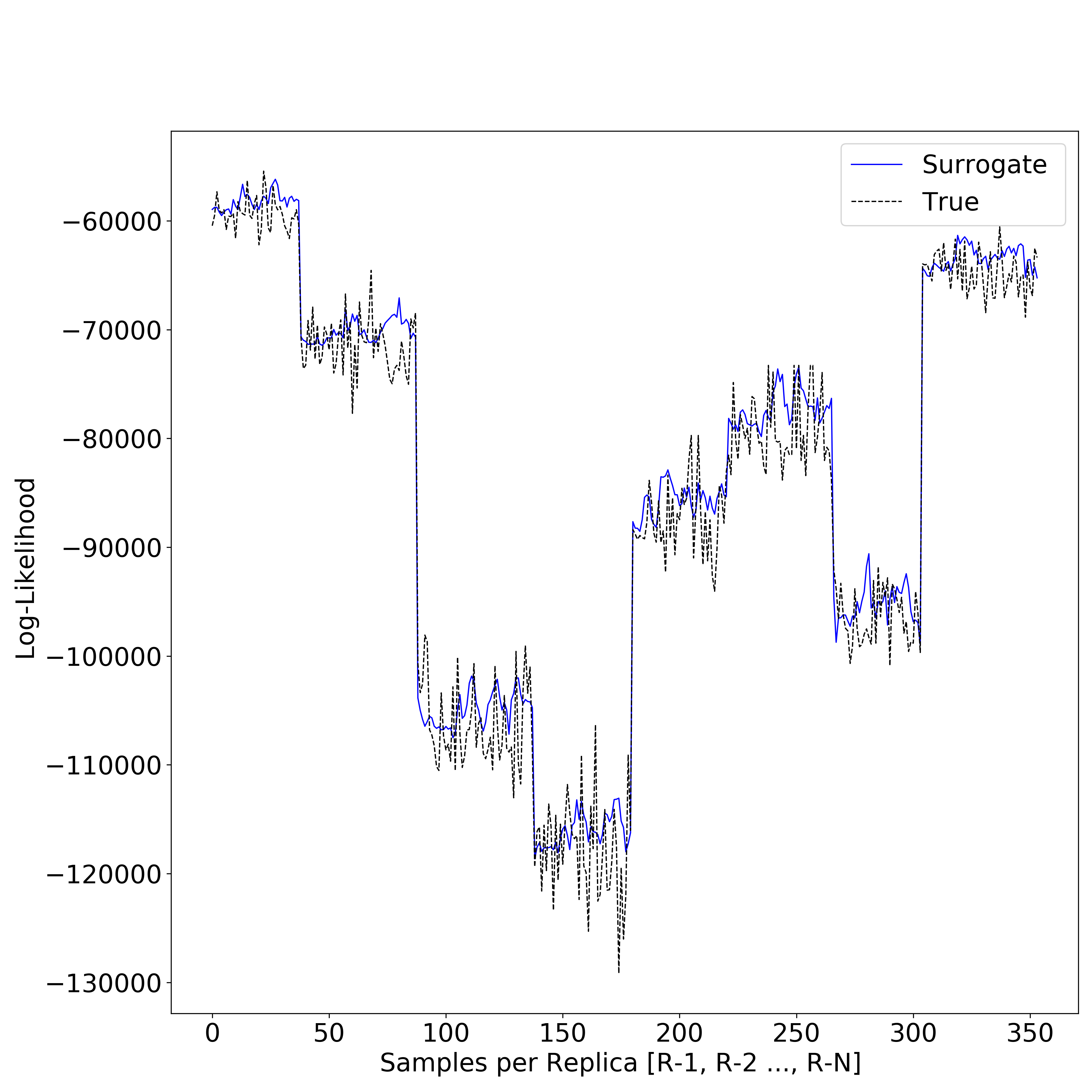}  
    \caption{Surrogate likelihood vs true likelihood estimation   for Continental-Margin problem ($RMSE_{sur} = 3605$). }
 \label{fig:surrogate_cm}
  \end{center}
\end{figure*}

\begin{figure*}[htbp!]
  \begin{center}
  \includegraphics[width=120mm]{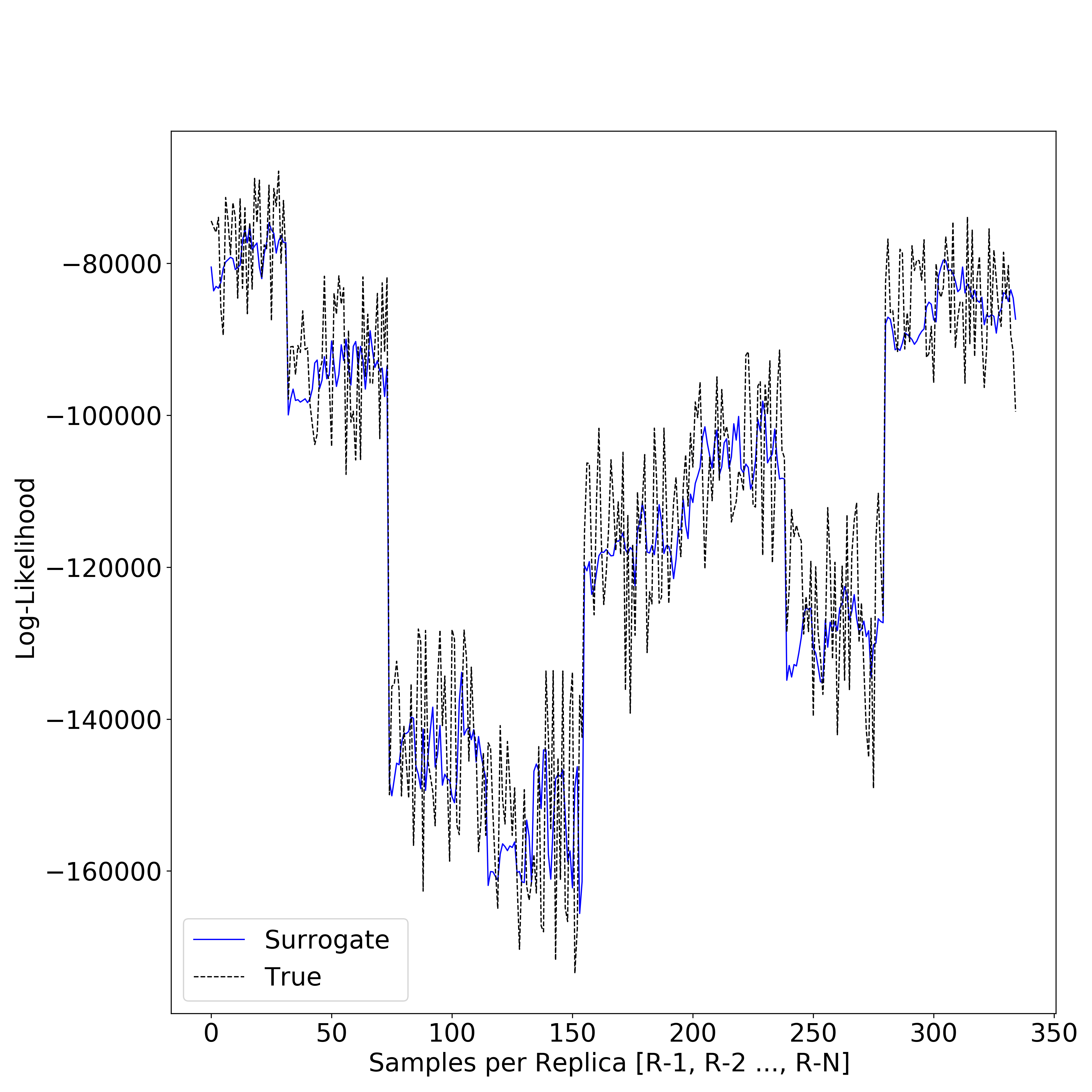}
    \caption{Surrogate likelihood vs true likelihood estimation   for Synthetic-Mountain problem ($RMSE_{sur} = 9917$). }
 \label{fig:surrogate_mn}
  \end{center}
\end{figure*}

Figure \ref{fig:surrogate_cm} and Figure \ref{fig:surrogate_mn} show the true likelihood and prediction by the surrogate for the Continental-Margin and Synthetic-Mountain problems, respectively. We notice that at certain intervals given in Figure \ref{fig:surrogate_cm}, given by different replica, there is inconsistency in the predictions. Moreover, Figure \ref{fig:surrogate_mn} shows that the log-likelihood is very chaotic, and hence there is difficulty in providing robust prediction at certain points in the time given by samples for the respective replica.

\begin{table*}
\small
\centering
\caption{Evaluation   for Continental-Margin problem}
\begin{tabular}{   l l l c c c l l}
\hline \hline 
	 Method & $S_{prob}$ & $\psi$ & $RMSE_{elev}$  & $RMSE_{elev}$  & $RMSE_{sed}$  & $RMSE_{sed}$  & time (s)  \\ 
	   &  &   & (mean) &   (std) &  (mean) &   (std)  &   \\
	 \hline \hline
  PT-Bayeslands & N/A & N/A & 78.80 &  10.03 & 35.91 &  11.36 & 3243.30 \\
    \hline 
     
SAPT-Bayeslands & 0.20 & 0.05&  75.53 &  9.89  &  35.68  & 10.93  &    3082.53  \\
SAPT-Bayeslands & 0.40 & 0.05& 80.22 & 15.63  & 44.72 & 16.52  & 2450.77  \\
SAPT-Bayeslands & 0.60 & 0.05& 82.04 & 8.23 & 44.33  & 13.37   &  1859.52   \\
SAPT-Bayeslands & 0.80 & 0.05& 79.30 & 26.70 & 43.29 & 18.68  & 1149.63   \\ 
     
	 \hline 
	  
SAPT-Bayeslands & 0.20 & 0.10 &  76.92 & 11.59  & 48.19   & 11.46     &    3075.31   \\
SAPT-Bayeslands & 0.40 & 0.10 & 82.43  & 11.58  & 46.47  & 12.55    & 2494.13   \\
SAPT-Bayeslands & 0.60 & 0.10 &  80.12 & 12.08  & 47.80  & 19.05  &  1934.34    \\
SAPT-Bayeslands & 0.80 & 0.10 &   88.81  & 20.61  & 51.12 &  14.26   & 1148.80    \\

	 \hline

SAPT-Bayeslands & 0.20 & 0.15 &   44.90 &  33.54 & 23.95 &    19.86      &    2914.06 \\
SAPT-Bayeslands & 0.40 & 0.15 & 73.64 & 8.05 & 38.53 &10.02     & 2495.56  \\
SAPT-Bayeslands & 0.60 & 0.15 &  83.38 & 8.45  & 51.15 &  19.07 &  1986.51   \\
SAPT-Bayeslands & 0.80 & 0.15 &   84.73 & 10.04 & 39.78 & 14.44   & 1294.64   \\ 
     
	 \hline
\end{tabular}

\label{tab:surrogatecmargin}
\end{table*}

\begin{table*}
\small
\centering
\caption{Performance comparison for respective problems and methods}
\begin{tabular}{ l l l l c c c l l }
\hline \hline
 Problem &Method & $S_{prob}$ & $\psi$ & $RMSE_{elev}$  & $RMSE_{elev}$  & $RMSE_{sed}$  & $RMSE_{sed}$  & time (s)  \\ 
	 &  &  &   & (mean) &   (std) &  (mean) &   (std)  &   \\
	 \hline \hline 
	 Continental Margin & PT-Bayeslands & N/A & N/A &78.80 &  10.03 & 35.91 &  11.36 & 3243.30 \\
	  
     & SAPT-Bayeslands &  0.60 &   0.05 &  82.0 & 8.23 & 44.33 & 13.37   & 1859.52  \\ 
	 \hline
	 Synthetic-Mountain & PT-Bayeslands & N/A & N/A &  106.10 &  48.24&  20.34  & 24.02 &   8474.67   \\
   
     & SAPT-Bayeslands &  0.60  & 0.05  & 104.88 &  5.51 & 11.87 &  8.69 &  4161.43\\  
     
     \hline
Tasmania & PT-Bayeslands & N/A & N/A &172.64&  10.74   & 3.90  & 0.50  & 
600293.61   \\ 
     & SAPT-Bayeslands & 0.60 & 0.05 &  179.67 & 19.71 & 3.91 & 0.10  & 
221942.41 \\ 
     \hline \hline
\end{tabular}

\label{tab:restasmania}
\end{table*}

\subsection{Discussion}

We observe that the surrogate probability is directly related to the computational performance; this is obvious since computational time depends on how often we use the surrogate. Our concern is about the prediction performance, especially while increasing the use of the surrogate as it could lower the accuracy, which can result in a poor estimation of the parameters. According to the results, the accuracy is well retained   given a higher probability  of using surrogates.  In the cross-section presented in the results for Continental-Margin and Synthetic Mountain problems, we find that there is not much difference in the accuracy given in prediction by the SAPT-Bayeslands when compared to PT-Bayeslands. Moreover, the application to a more computationally intensive problem (Tasmania), we find that a significant reduction in computational time is achieved. Although we demonstrated the method using small-scale  models that run within a few seconds to minutes,  the computational costs of continental-scale Badlands models is extensive. For instance, the computational time for a  5-kilometre resolution for Australian continent Badlands model for 149 million years is about 72 hours; hence, in the case when thousands of samples are required,  the use of surrogates can be beneficial. We note that improved efficiency of the surrogate-assisted Bayeslands comes at the cost of  accuracy for some problems (in case of Tasmania problem), and there is a trade-off between accuracy and computational time. 

 In future work, rather than a global surrogate model, we could use the local surrogate model on its own, where the training only takes place in the local surrogates by relying on the history of the likelihood and hence taking a univariate time series prediction approach using neural networks. Our primary contribution is in terms of the parallel computing based open-source software and the proposed underlying framework for incorporating surrogates, taking into account complex issues such as inter-process communication. This opens the road to using different types of surrogate models while using the underlying framework and open-source software.  Given that the sediment erosion/deposition is temporal, other ways of formulating the likelihood could be possible, for instance, we could have a hierarchical Bayesian model  with two stages for MCMC sampling \citep{chib1999mcmc,wikle1998hierarchical}.

The initial evaluation for the setup surrogate model shows that it is best to use a  transfer learning approach where the knowledge from the past surrogate interval is utilised and refined with new surrogate data. This consumes much less time than accumulating data and training the surrogate from scratch at every surrogate interval. We note that in the case when we use the surrogate model for pseudo-likelihood, there is no prediction given by the surrogate model. The prediction (elevation topography and erosion-deposition) during sampling are gathered only from the true Badlands model evaluation rather than the surrogate. In this way, one could argue that the surrogate model is not mimicking the true model; however, we are guiding the sampling algorithm towards forming better proposals without evaluation of the true model. A direction forward is in incorporating other forms of surrogates, such as running low-resolution Badlands model as the surrogate  which would be computationally faster in evaluating the proposals; however,   limitations in terms of effect of resolution setting on Badlands topography simulation  may exist.

Furthermore, computationally efficient implementations of landscape evolution models that only feature landscape evolution \citep{BRAUN2013170} could be used as the surrogate, while we could use Badlands model that features both landscape evolution and erosion/deposition as the true model. We could also use computationally efficient implementations of landscape evolution models that consider parallel processing \citep{HASSAN2018263}   in the Bayeslands framework. In this case, the challenge would be in allocating specialised processing cores for Badlands and others for parallel tempering MCMC.

We adapted the surrogate framework developed for machine learning \citep{chandra2020SAPT} with a  different  proposal  distribution instead of using gradient-based proposals. Gradient-based   parameter estimation has been very popular in machine learning due to availability of gradient information.  Due to the complexity in geological or geophysical numerical forward models, it is challenging to obtain gradients which have been the case of Badlands, landscape evolution model. We used random-walk  and adaptive random-walk proposal distributions which have   limitations; hence, we need to incorporate advanced meta-heuristic techniques to form non-gradient based proposals for efficient search. Our study is limited to a relatively small seat of free parameters, and a significant challenge would be to develop surrogate models with an increased set of parameters.

\conclusions  

We presented a novel application of surrogate-assisted parallel tempering that features parallel computing for landscape evolution models using Badlands. Initially, we experimented with two different approaches for training the surrogate model, where we found that transfer learning-based approach is beneficial and could help reduce the computational time of the surrogate.  Using this approach, we presented the experiments that featured evaluating certain key parameters of the surrogate-based framework. In general, we observed that the proposed framework lowers the computational time significantly while maintaining the required quality in parameter estimation and uncertainty quantification. 

In future work, we envision to apply the proposed framework to more complex applications such as the evolution of continental-scale landscapes and basins over millions of years.  We could use the approach for other forward models such as those that feature geological reef development or lithospheric deformation. Furthermore, the posterior distribution of our parameters require multi-modal sampling methods; hence, a combination of meta-heuristics for proposals with surrogate assisted parallel tempering could improve exploration features and also help in lowering the computational costs.

\codeavailability{https://github.com/intelligentEarth/surrogateBayeslands}  

\codeavailability{https://doi.org/10.5281/zenodo.3892277}  
 \appendix

 \subsection{Parallel tempering MCMC}
  Parallel tempering   MCMC  features massive parallelism with enhanced  exploration capabilities.  It features several replicas with slight variations in the acceptance criteria through relaxation of the likelihood with a temperature ladder that affects the replica sampling acceptance criterion. The replicas associated with higher temperature levels have more chance in accepting weaker proposals   which could help in escaping a local minimum.  Given an ensemble of $M$ replicas  defined by a temperature ladder,   we define the state by   $X= {x_1,x_2, ..., x_M}$; where $x_i$ is the replica at temperature level $T_i$.  We construct a Markov chain  to sample proposal  $x_i$ and  evaluate it using the  likelihood $L(x_i)$ for  each replica defined by temperature level $T_i$. At each iteration, the Markov chain can feature two types of transitions that include the \textit{Metropolis transition} and the \textit{replica transition}. 

In the \textit{Metropolis transition} phase, we independently sample each replica   to perform local \textit{Monte Carlo} moves  as defined by the temperature ladder  for the replica by relaxing or changing the likelihood   in relation to the temperature level $L(x_i)/T_i$. We sample  configuration $x^*_i$ from a proposal distribution $q_i(.|x_i)$. The \textit{Metropolis-Hastings} ratio at temperature level  $T_i$ is given asby

\begin{equation}
L_{local}(x_i \rightarrow x^*_i) =  exp(- \frac{1}{T_i}   (L(x^*_i) - L(x_i)))  
\end{equation}

where $L$  represents the likelihood at the $local$ replica. We accept the new state  with probability, $min(1, L_{local}(x_i \rightarrow x^*_i))$. The detailed balance condition holds for each MCMC replica;  therefore, it holds for the ensemble system    \citep{Calderhead17408}. 
 
In the \textit{replica transition} phase, we  consider the exchange of the current state between two neighbouring replicas based on the Metropolis-Hasting acceptance criteria.  Hence,  given a probability $\alpha$,   we exchange a  pair of replica defined by two neighbouring temperature levels, $T_i$ and $T_{i+1}$.

\begin{equation}
 x_i \leftrightarrow x_{i+1}  
\end{equation}

 The exchange of neighbouring replicas provide an efficient balance between local and global exploration \citep{sambridge2013parallel}. The temperature ladder and replica-exchange have been of the focus of investigation in the past \citep{calvo2005all,Liu13749,bittner2008make,patriksson2008temperature} and there is a consensus that they need to be tailored for different types of problems given by their likelihood landscape.   In this paper, the selection of temperature spacing between the replicas is carried out using a  Geometric spacing methodology   \citep{vousden2015dynamic}, given as follows 

\begin{equation}
 T_i = T_{max}^{(i-1)/(M-1)}
 \label{eq:geometric}
 \end{equation}
 
  where $i = 1, \ldots, M$ and $T_{max}$ is maximum temperature which is user defined and dependent on the problem.

  \subsection{ Training the neural network surrogate model}
  
 We note that stochastic gradient descent maintains a single learning rate for all weight updates and typically the learning rate does not change during the training. 
Adam (adaptive moment estimation) learning algorithm \cite{kingma2014adam} differs from classical stochastic gradient descent, as the learning rate is maintained for each network weight and separately adapted as learning unfolds. Adam computes individual adaptive learning rates for different parameters from estimates of first and second moments of the gradients. Adam features the strengths of  \textit{root mean square propagation}  and \textit{adaptive gradient algorithm} (AdaGrad) \citep{kingma2014adam,duchi2011adaptive}.  Adam has shown better results when compared to stochastic gradient descent,  RMSprop and AdaGrad. Hence, we consider Adam as the designated algorithm for the neural network-based surrogate model.  We formulate the learning procedure through weight update for  iteration number $t$ for weights $\bf{W}$ and biases $\bf{b}$, by 

\begin{eqnarray}
\Theta_{t-1}&=&[\bf{W_{t-1},b_{t-1}}]\nonumber\\
g_t &=& \nabla_{\Theta}J_t(\Theta_{t-1})\nonumber\\
m_t &=& \beta_1.m_{t-1} + (1-\beta_1).g_t\nonumber\\
v_t &=& \beta_2.v_{t-1} + (1-\beta_2).g_t^2\nonumber\\
\hat{m}_t &=& m_t/(1-\beta^t_1)\nonumber\\
\hat{v}_t &=& v_t/(1-\beta^t_2)\nonumber\\
\Theta_t &=& \Theta_{t-1} -  \alpha.\hat{m}_t/(\sqrt[]{\hat{v}_t} + \epsilon)
\end{eqnarray}

 where $m_t, v_t$ are the respective first  and second  moment vectors for iteration $t$; $\beta_1, \beta_2$ are constants $\in [0,1]$, $\alpha$ is the learning rate, and $\epsilon$ is a close to zero constant.





\appendixfigures  

\appendixtables   


\authorcontribution{R. Chandra led the project and contributed to writing the paper and designing the experiments. D. Azam contributed by running experiments and providing documentation of the results. A. Kapoor contributed in terms of programming, running experiments and providing documentation of the results. R. D. Müller contributed by managing the project,   writing the paper, and providing analysis of the results. } 

\competinginterests{No competing interests are present.} 


\begin{acknowledgements}

 We would like to thank Konark Jain for technical support. 

\end{acknowledgements}







\bibliographystyle{copernicus} 
\bibliography{bibliography,2018,Bays,Chandra-Rohitash,surrogate,extra} 

\begin{thebibliography}{61}
\providecommand{\natexlab}[1]{#1}
\providecommand{\url}[1]{{\tt #1}}
\providecommand{\urlprefix}{URL }
\expandafter\ifx\csname urlstyle\endcsname\relax
  \providecommand{\doi}[1]{https://doi.org/\discretionary{}{}{}#1}\else
  \providecommand{\doi}{https://doi.org/\discretionary{}{}{}\begingroup
  \urlstyle{rm}\Url}\fi

\bibitem[{Adams et~al.(2017)Adams, Gasparini, Hobley, Tucker, Hutton,
  Nudurupati, and Istanbulluoglu}]{Adams2017}
Adams, J.~M., Gasparini, N.~M., Hobley, D. E.~J., Tucker, G.~E., Hutton, E.
  W.~H., Nudurupati, S.~S., and Istanbulluoglu, E.: The Landlab v1.0
  OverlandFlow component: a Python tool for computing shallow-water flow across
  watersheds, Geoscientific Model Development, 10, 1645--1663, 2017.

\bibitem[{Ampomah et~al.(2017)Ampomah, Balch, Will, Cather, Gunda, and
  Dai}]{ampomah2017co}
Ampomah, W., Balch, R., Will, R., Cather, M., Gunda, D., and Dai, Z.:
  Co-optimization of CO2-EOR and Storage Processes under Geological
  Uncertainty, Energy Procedia, 114, 6928--6941, 2017.

\bibitem[{Asher et~al.(2015)Asher, Croke, Jakeman, and
  Peeters}]{asher2015review}
Asher, M.~J., Croke, B.~F., Jakeman, A.~J., and Peeters, L.~J.: A review of
  surrogate models and their application to groundwater modeling, Water
  Resources Research, 51, 5957--5973, 2015.

\bibitem[{Bittner et~al.(2008)Bittner, Nu{\ss}baumer, and
  Janke}]{bittner2008make}
Bittner, E., Nu{\ss}baumer, A., and Janke, W.: Make life simple: Unleash the
  full power of the parallel tempering algorithm, Physical review letters, 101,
  130\,603, 2008.

\bibitem[{Bottou(1991)}]{bottou1991stochastic}
Bottou, L.: Stochastic gradient learning in neural networks, Proceedings of
  Neuro-N{\i}mes, 91, 12, 1991.

\bibitem[{Bottou(2010)}]{bottou2010large}
Bottou, L.: Large-scale machine learning with stochastic gradient descent, in:
  Proceedings of COMPSTAT'2010, pp. 177--186, Springer, 2010.

\bibitem[{Braun and Willett(2013)}]{BRAUN2013170}
Braun, J. and Willett, S.~D.: A very efficient O(n), implicit and parallel
  method to solve the stream power equation governing fluvial incision and
  landscape evolution, Geomorphology, 180-181, 170 -- 179, 2013.

\bibitem[{Broomhead and Lowe(1988)}]{broomhead1988radial}
Broomhead, D.~S. and Lowe, D.: Radial basis functions, multi-variable
  functional interpolation and adaptive networks, Tech. rep., Royal Signals and
  Radar Establishment Malvern (United Kingdom), 1988.

\bibitem[{Calderhead(2014)}]{Calderhead17408}
Calderhead, B.: A general construction for parallelizing Metropolis-Hastings
  algorithms, Proceedings of the National Academy of Sciences, 111,
  17\,408--17\,413,   2014.

\bibitem[{Calvo(2005)}]{calvo2005all}
Calvo, F.: All-exchanges parallel tempering, The Journal of chemical physics,
  123, 124\,106, 2005.

\bibitem[{Campforts et~al.(2017)Campforts, Schwanghart, and
  Govers}]{Campforts2017}
Campforts, B., Schwanghart, W., and Govers, G.: Accurate simulation of
  transient landscape evolution by eliminating numerical diffusion: the
  TTLEM~1.0 model, Earth Surface Dynamics, 5, 47--66, 2017.

\bibitem[{Chandra et~al.(2019{\natexlab{a}})Chandra, Azam, M{\"u}ller, Salles,
  and Cripps}]{Chandra2018_Bayeslands_}
Chandra, R., Azam, D., M{\"u}ller, R.~D., Salles, T., and Cripps, S.:
  BayesLands: A {Bayesian} inference approach for parameter uncertainty
  quantification in Badlands, Computers \& Geosciences, 131, 89--101,
  2019{\natexlab{a}}.

\bibitem[{Chandra et~al.(2019{\natexlab{b}})Chandra, Jain, Deo, and
  Cripps}]{chandra2019langevin}
Chandra, R., Jain, K., Deo, R.~V., and Cripps, S.: Langevin-gradient parallel
  tempering for {Bayesian} neural learning, Neurocomputing, 359, 315 -- 326,
   2019{\natexlab{b}}.

\bibitem[{Chandra et~al.(2019{\natexlab{c}})Chandra, M{\"u}ller, Azam, Deo,
  Butterworth, Salles, and Cripps}]{chandra2018PT-Bayes_}
Chandra, R., M{\"u}ller, R.~D., Azam, D., Deo, R., Butterworth, N., Salles, T.,
  and Cripps, S.: Multi-core parallel tempering {Bayeslands} for basin and
  landscape evolution, Geochemistry, Geophysics, Geosystems, 20, 5082--5104, 2019{\natexlab{c}}.

\bibitem[{Chandra et~al.(2020)Chandra, Jain, Arpit, and
  Ashray}]{chandra2020SAPT}
Chandra, R., Jain, K., Arpit, K., and Ashray, A.: Surrogate-assisted parallel
  tempering for {Bayesian} neural learning, Engineering Applications of
  Artificial Intelligence, 94, 103\,700, \doi{10.1016/j.engappai.2020.103700},
  2020.

\bibitem[{Chib and Carlin(1999)}]{chib1999mcmc}
Chib, S. and Carlin, B.~P.: On MCMC sampling in hierarchical longitudinal
  models, Statistics and Computing, 9, 17--26, 1999.

\bibitem[{Chollet et~al.(2015)}]{chollet2015keras}
Chollet, F. et~al.: Keras, \url{https://keras.io}, 2015.

\bibitem[{D{\'\i}az-Manr{\'\i}quez et~al.(2016)D{\'\i}az-Manr{\'\i}quez,
  Toscano, Barron-Zambrano, and Tello-Leal}]{diaz2016review}
D{\'\i}az-Manr{\'\i}quez, A., Toscano, G., Barron-Zambrano, J.~H., and
  Tello-Leal, E.: A review of surrogate assisted multiobjective evolutionary
  algorithms, Computational intelligence and neuroscience, 2016, 2016.

\bibitem[{Duchi et~al.(2011)Duchi, Hazan, and Singer}]{duchi2011adaptive}
Duchi, J., Hazan, E., and Singer, Y.: Adaptive subgradient methods for online
  learning and stochastic optimization, Journal of Machine Learning Research,
  12, 2121--2159, 2011.

\bibitem[{Gelman et~al.(1992)Gelman, Rubin et~al.}]{gelman1992inference}
Gelman, A., Rubin, D.~B., et~al.: Inference from iterative simulation using
  multiple sequences, Statistical science, 7, 457--472, 1992.

\bibitem[{Geyer and Thompson(1995)}]{geyer1995annealing}
Geyer, C.~J. and Thompson, E.~A.: Annealing Markov chain Monte Carlo with
  applications to ancestral inference, Journal of the American Statistical
  Association, 90, 909--920, 1995.

\bibitem[{Haario et~al.(2001)Haario, Saksman, Tamminen
  et~al.}]{haario2001adaptive}
Haario, H., Saksman, E., Tamminen, J., et~al.: An adaptive Metropolis
  algorithm, Bernoulli, 7, 223--242, 2001.

\bibitem[{Hassan et~al.(2018)Hassan, Gurnis, Williams, and
  Müller}]{HASSAN2018263}
Hassan, R., Gurnis, M., Williams, S.~E., and Müller, R.~D.: SPGM: A Scalable
  PaleoGeomorphology Model, SoftwareX, 7, 263 -- 272, 2018.

\bibitem[{Hastings(1970)}]{hastings1970monte}
Hastings, W.~K.: Monte Carlo sampling methods using Markov chains and their
  applications, Biometrika, 57, 97--109, 1970.

\bibitem[{Hobley et~al.(2011)Hobley, Sinclair, Mudd, and Cowie}]{Hobley2011}
Hobley, D. E.~J., Sinclair, H.~D., Mudd, S.~M., and Cowie, P.~A.: Field
  calibration of sediment flux dependent river incision, Journal of Geophysical
  Research: Earth Surface, 116, 2011.

\bibitem[{Howard et~al.(1994)Howard, Dietrich, and Seidl}]{Howard1994}
Howard, A.~D., Dietrich, W.~E., and Seidl, M.~A.: Modeling fluvial erosion on
  regional to continental scales, Journal of Geophysical Research: Solid Earth,
  99, 13\,971--13\,986, 1994.

\bibitem[{Hukushima and Nemoto(1996)}]{hukushima1996exchange}
Hukushima, K. and Nemoto, K.: Exchange Monte Carlo method and application to
  spin glass simulations, Journal of the Physical Society of Japan, 65,
  1604--1608, 1996.

\bibitem[{Jin(2011)}]{jin2011surrogate}
Jin, Y.: Surrogate-assisted evolutionary computation: Recent advances and
  future challenges, Swarm and Evolutionary Computation, 1, 61--70, 2011.

\bibitem[{Kingma and Ba(2014)}]{kingma2014adam}
Kingma, D.~P. and Ba, J.: Adam: A method for stochastic optimization, arXiv
  preprint arXiv:1412.6980, 2014.

\bibitem[{Lamport(1986)}]{lamport1986interprocess}
Lamport, L.: On interprocess communication, Distributed computing, 1, 86--101,
  1986.

\bibitem[{Letsinger et~al.(1996)Letsinger, Myers, and
  Lentner}]{letsinger1996response}
Letsinger, J.~D., Myers, R.~H., and Lentner, M.: Response surface methods for
  bi-randomization structures, Journal of quality technology, 28, 381--397,
  1996.

\bibitem[{Lim et~al.(2010)Lim, Jin, Ong, and Sendhoff}]{lim2010generalizing}
Lim, D., Jin, Y., Ong, Y.-S., and Sendhoff, B.: Generalizing surrogate-assisted
  evolutionary computation, IEEE Transactions on Evolutionary Computation, 14,
  329--355, 2010.

\bibitem[{Liu et~al.(2005)Liu, Kim, Friesner, and Berne}]{Liu13749}
Liu, P., Kim, B., Friesner, R.~A., and Berne, B.~J.: Replica exchange with
  solute tempering: A method for sampling biological systems in explicit water,
  Proceedings of the National Academy of Sciences, 102, 13\,749--13\,754, 2005.

\bibitem[{Marinari and Parisi(1992)}]{marinari1992simulated}
Marinari, E. and Parisi, G.: Simulated tempering: a new Monte Carlo scheme, EPL
  (Europhysics Letters), 19, 451, 1992.

\bibitem[{Metropolis et~al.(1953)Metropolis, Rosenbluth, Rosenbluth, Teller,
  and Teller}]{metropolis1953equation}
Metropolis, N., Rosenbluth, A.~W., Rosenbluth, M.~N., Teller, A.~H., and
  Teller, E.: Equation of state calculations by fast computing machines, The
  journal of chemical physics, 21, 1087--1092, 1953.

\bibitem[{Montgomery and Vernon M.~Bettencourt(1977)}]{Douglas1977}
Montgomery, D.~C. and Vernon M.~Bettencourt, J.: Multiple response surface
  methods in computer simulation, Simulation, 29, 113--121, 1977.

\bibitem[{Navarro et~al.(2018)Navarro, Le~Ma{\^\i}tre, Hoteit, George, Mandli,
  and Knio}]{navarro2018surrogate}
Navarro, M., Le~Ma{\^\i}tre, O.~P., Hoteit, I., George, D.~L., Mandli, K.~T.,
  and Knio, O.~M.: Surrogate-based parameter inference in debris flow model,
  Computational Geosciences, pp. 1--17, 2018.

\bibitem[{Neal(1996)}]{neal1996sampling}
Neal, R.~M.: Sampling from multimodal distributions using tempered transitions,
  Statistics and computing, 6, 353--366, 1996.

\bibitem[{Neal(2012)}]{neal2012bayesian}
Neal, R.~M.: Bayesian learning for neural networks, vol. 118, Springer Science
  \& Business Media, 2012.

\bibitem[{Olierook et~al.(2020)Olierook, Scalzo, Kohn, Chandra, Farahbakhsh,
  Clark, Reddy, and Müller}]{OLIEROOK2020}
Olierook, H.~K., Scalzo, R., Kohn, D., Chandra, R., Farahbakhsh, E., Clark, C.,
  Reddy, S.~M., and Müller, R.~D.: Bayesian geological and geophysical data
  fusion for the construction and uncertainty quantification of {3D} geological
  models, Geoscience Frontiers,
  \doi{https://doi.org/10.1016/j.gsf.2020.04.015}, 2020.

\bibitem[{Ong et~al.(2003)Ong, Nair, and Keane}]{ong2003evolutionary}
Ong, Y.~S., Nair, P.~B., and Keane, A.~J.: Evolutionary optimization of
  computationally expensive problems via surrogate modeling, AIAA journal, 41,
  687--696, 2003.

\bibitem[{Ong et~al.(2005)Ong, Nair, Keane, and Wong}]{ong2005surrogate}
Ong, Y.~S., Nair, P., Keane, A., and Wong, K.: Surrogate-assisted evolutionary
  optimization frameworks for high-fidelity engineering design problems, in:
  Knowledge Incorporation in Evolutionary Computation, pp. 307--331, Springer,
  2005.

\bibitem[{Pall et~al.(2020)Pall, Chandra, Azam, Salles, Webster, Scalzo, and
  Cripps}]{JPall_BayesReef2020}
Pall, J., Chandra, R., Azam, D., Salles, T., Webster, J.~M., Scalzo, R., and
  Cripps, S.: Bayesreef: A {Bayesian} inference framework for modelling reef
  growth in response to environmental change and biological dynamics,
  Environmental Modelling \& Software, p. 104610, 2020.

\bibitem[{Patriksson and van~der Spoel(2008)}]{patriksson2008temperature}
Patriksson, A. and van~der Spoel, D.: A temperature predictor for parallel
  tempering simulations, Physical Chemistry Chemical Physics, 10, 2073--2077,
  2008.

\bibitem[{Raftery and Lewis(1996)}]{raftery1996}
Raftery, A.~E. and Lewis, S.~M.: Implementing mcmc, Markov chain Monte Carlo in
  practice, pp. 115--130, 1996.

\bibitem[{Rasmussen(2004)}]{rasmussen2004gaussian}
Rasmussen, C.~E.: Gaussian processes in machine learning, in: Advanced lectures
  on machine learning, pp. 63--71, Springer, 2004.

\bibitem[{Razavi et~al.(2012)Razavi, Tolson, and Burn}]{razavi2012review}
Razavi, S., Tolson, B.~A., and Burn, D.~H.: Review of surrogate modeling in
  water resources, Water Resources Research, 48, 2012.

\bibitem[{Salles and Hardiman(2016)}]{salles2016badlands}
Salles, T. and Hardiman, L.: Badlands: An open-source, flexible and parallel
  framework to study landscape dynamics, Computers \& Geosciences, 91, 77--89,
  2016.

\bibitem[{Salles et~al.(2018)Salles, Ding, and Brocard}]{salles2018pybadlands}
Salles, T., Ding, X., and Brocard, G.: pyBadlands: A framework to simulate
  sediment transport, landscape dynamics and basin stratigraphic evolution
  through space and time, PloS one, 13, e0195\,557, 2018.

\bibitem[{Sambridge(1999)}]{sambridge1999geophysical}
Sambridge, M.: Geophysical inversion with a neighbourhood algorithm—II.
  Appraising the ensemble, Geophysical Journal International, 138, 727--746,
  1999.

\bibitem[{Sambridge(2013)}]{sambridge2013parallel}
Sambridge, M.: A parallel tempering algorithm for probabilistic sampling and
  multimodal optimization, Geophysical Journal International, 196, 357--374,
  2013.

\bibitem[{Scalzo et~al.(2019)Scalzo, Kohn, Olierook, Houseman, Chandra,
  Girolami, and Cripps}]{scalzo2019efficiency}
Scalzo, R., Kohn, D., Olierook, H., Houseman, G., Chandra, R., Girolami, M.,
  and Cripps, S.: Efficiency and robustness in Monte Carlo sampling for 3-D
  geophysical inversions with Obsidian v0. 1.2: setting up for success,
  Geoscientific Model Development, 12, 2941--2960, 2019.

\bibitem[{Scher(2018)}]{Scher2018}
Scher, S.: Toward Data-Driven Weather and Climate Forecasting: Approximating a
  Simple General Circulation Model With Deep Learning, Geophysical Research
  Letters, 45, 1--7, \doi{10.1029/2018GL080704},
  \urlprefix\url{https://agupubs.onlinelibrary.wiley.com/doi/abs/10.1029/2018GL080704},
  2018.

\bibitem[{Tandjiria et~al.(2000)Tandjiria, Teh, and
  Low}]{tandjiria2000reliability}
Tandjiria, V., Teh, C.~I., and Low, B.~K.: Reliability analysis of laterally
  loaded piles using response surface methods, Structural Safety, 22, 335--355,
  2000.

\bibitem[{Tucker and Hancock(2010)}]{Tucker10}
Tucker, G.~E. and Hancock, G.~R.: Modelling landscape evolution, Earth Surface
  Processes and Landforms, 35, 28--50, 2010.

\bibitem[{van~der Merwe et~al.(2007)van~der Merwe, Leen, Lu, Frolov, and
  Baptista}]{van2007fast}
van~der Merwe, R., Leen, T.~K., Lu, Z., Frolov, S., and Baptista, A.~M.: Fast
  neural network surrogates for very high dimensional physics-based models in
  computational oceanography, Neural Networks, 20, 462--478, 2007.

\bibitem[{van Ravenzwaaij et~al.(2016)van Ravenzwaaij, Cassey, and
  Brown}]{van2016simple}
van Ravenzwaaij, D., Cassey, P., and Brown, S.~D.: A simple introduction to
  Markov Chain Monte--Carlo sampling, Psychonomic bulletin \& review, pp.
  1--12, 2016.

\bibitem[{Vousden et~al.(2015)Vousden, Farr, and Mandel}]{vousden2015dynamic}
Vousden, W., Farr, W.~M., and Mandel, I.: Dynamic temperature selection for
  parallel tempering in Markov chain Monte Carlo simulations, Monthly Notices
  of the Royal Astronomical Society, 455, 1919--1937, 2015.

\bibitem[{Whipple and Tucker(2002)}]{Whipple2002}
Whipple, K.~X. and Tucker, G.~E.: Implications of sediment-flux-dependent river
  incision models for landscape evolution, Journal of Geophysical Research:
  Solid Earth, 107, 1--20, 2002.

\bibitem[{Wikle et~al.(1998)Wikle, Berliner, and
  Cressie}]{wikle1998hierarchical}
Wikle, C.~K., Berliner, L.~M., and Cressie, N.: Hierarchical {Bayesian}
  space-time models, Environmental and Ecological Statistics, 5, 117--154,
  1998.

\bibitem[{Zhou et~al.(2007)Zhou, Ong, Nair, Keane, and Lum}]{zhou2007combining}
Zhou, Z., Ong, Y.~S., Nair, P.~B., Keane, A.~J., and Lum, K.~Y.: Combining
  global and local surrogate models to accelerate evolutionary optimization,
  IEEE Transactions on Systems, Man, and Cybernetics, Part C (Applications and
  Reviews), 37, 66--76, 2007.

\end{thebibliography}


%

\end{document}